\documentclass{article}
\PassOptionsToPackage{numbers}{natbib}
\usepackage[preprint]{neurips_data_2024}

\usepackage{graphicx} 
\usepackage{microtype}
\usepackage{hyperref}
\usepackage{url}
\usepackage{bbold}
\usepackage{booktabs}
\usepackage{xcolor}
\usepackage{tcolorbox}
\usepackage{xspace}
\usepackage{amssymb}
\usepackage{enumerate}
\usepackage{multirow}
\usepackage{lipsum} % For generating dummy text
\usepackage{booktabs} % For better table formatting
\usepackage{subcaption} % For subfigures & subtables
\usepackage{tabularx} % For better table formatting

\usepackage{pifont}% http://ctan.org/pkg/pifont
\usepackage{float}
\usepackage{caption}

\newcommand{\cmark}{\ding{51}}%

\definecolor{LightGray}{gray}{0.9}
\definecolor{pastelblue}{RGB}{173,216,230}
\definecolor{pastelpink}{RGB}{255,182,193}
\definecolor{pastelgreen}{RGB}{152,251,152}

\newtcolorbox{pastelbox}[2][]{colback=#1!10!white, colframe=#1!80!black, 
    boxrule=0.5mm, 
    arc=1.5mm, 
    auto outer arc,
    left=1mm,
    right=1mm,
    top=1mm,
    bottom=1mm,
    title=#2}

\usepackage{lipsum}
\usepackage{wrapfig}
\usepackage{graphicx,multirow}
\usepackage{amsmath}
\usepackage{booktabs, multirow} % for borders and merged ranges
\usepackage{soul}% for underlines
\usepackage{changepage,threeparttable} % for wide tables

\newdimen\abovecrulesep
\newdimen\belowcrulesep
\makeatletter
\patchcmd{\@@@cmidrule}{\aboverulesep}{\abovecrulesep}{}{}
\patchcmd{\@xcmidrule}{\belowrulesep}{\belowcrulesep}{}{}

\definecolor{demphcolor}{RGB}{144, 144, 144}
\definecolor{mygray}{gray}{0.4}
\definecolor{lightgray}{rgb}{0.9, 0.9, 0.9}
\newcommand{\demph}[1]{\textcolor{demphcolor}{#1}}

\newlength\savewidth
\newcommand\shline{\noalign{\global\savewidth\arrayrulewidth\global\arrayrulewidth 1pt}\hline\noalign{\global\arrayrulewidth\savewidth}}

\newcommand{\tablestyle}[2]{\setlength{\tabcolsep}{#1}\renewcommand{\arraystretch}{#2}\centering\footnotesize}
\makeatletter\renewcommand\paragraph{\@startsection{paragraph}{4}{\z@}{0em\@plus1ex\@minus.2ex}{0em}{\normalfont\normalsize\bfseries}}
\makeatother

\newcolumntype{C}[1]{>{\centering\arraybackslash}p{#1}}
\newcolumntype{R}[1]{>{\raggedleft\arraybackslash}p{#1}}
\newcolumntype{L}[1]{>{\raggedright\arraybackslash}p{#1}}

\preto\align{\small}
\expandafter\preto\csname align*\endcsname{\small}
\preto\equation{\par\nobreak\small\noindent}

\usepackage{enumitem}
\usepackage[font=footnotesize,labelfont=bf,aboveskip=3pt,belowskip=-10pt]{caption}
\usepackage{amsmath}
\usepackage{amssymb}
\usepackage{longtable}
\usepackage{tcolorbox}
\tcbuselibrary{skins,xparse,breakable}
\definecolor{salmon}{rgb}{0.98, 0.5, 0.45}

\newcommand{\ourdata}[0]{\textsc{CertainlyUncertain}\xspace}

\title{%\includegraphics[width=1.5em, trim=3cm 3cm 3cm 2.5cm, clip]{figures/sc.pdf} 
Certainly Uncertain: A Benchmark and Metric \\ for Multimodal Epistemic and Aleatoric Awareness
}

\DeclareSymbolFont{extraup}{U}{zavm}{m}{n}
\DeclareMathSymbol{\varheart}{\mathalpha}{extraup}{86}
\DeclareMathSymbol{\vardiamond}{\mathalpha}{extraup}{87}

\author{ 
  Khyathi Raghavi Chandu$^\clubsuit$  $\thanks{Equal contribution}$\quad
  Linjie Li$^\vardiamond$$^\spadesuit$ $^*$ \quad \\
  \textbf{Anas Awadalla}$^\vardiamond$\quad 
  \textbf{Ximing Lu}$^\vardiamond$\quad
  \textbf{Jae Sung Park}$^\vardiamond$ \quad
  \textbf{Jack Hessel}$^\varheart$ \quad \\
  \textbf{Lijuan Wang}$^\spadesuit$ \quad
  \textbf{Yejin Choi}$^\clubsuit$ $^\vardiamond$ \quad
  ~\vspace{0.4em}\\ 
  $^\clubsuit$Allen Institute for AI \qquad
  $^\vardiamond$University of Washington \qquad
  $^\varheart$Samaya AI \qquad
  $^\spadesuit$Microsoft 
  \\ 
}

\begin{document}

\maketitle

\begin{abstract}
% \textbf{Option 1: bottom up}
% Vision-language models (VLMs) currently lack self-awareness of their limitations and the ability to acknowledge "I don't know" when faced with ambiguous or unanswerable queries which is crucial for ensuring reliability and trustworthiness. The lack of this epistemic (due to insufficient information) and aleatoric (due to inherent unpredictability) awareness leads to overconfident and potentially misleading responses. This gap is exacerbated by the absence of datasets that address unanswerable questions. To bridge this gap, we propose a comprehensive framework of data creation, modeling techniques and evaluation methods to improve different fine-grained types of unanswerability. Leveraging this framework, we develop a novel dataset comprising zzz instances specifically designed to capture various dimensions of unanswerability. We conduct extensive evaluations of different training techniques on our dataset, alongside a few existing datasets, to benchmark performance improvements. Our findings demonstrate that incorporating our unanswerability-focused dataset significantly enhances the models' ability to handle uncertain scenarios, increasing their robustness and fostering user trust by making them more self-aware of their limitations and confidence levels.

% TODO: Name for the dataset
% TODO: which technique
% TODO: rephrase to top-down
% %UncertainVQA

The ability to acknowledge the inevitable uncertainty in their knowledge and reasoning is a prerequisite for AI systems to be truly truthful and reliable.
% In this paper, we study how multimodal large language models (MLLMs) perform when faced with uncertain scenarios. 
In this paper, we present a taxonomy of uncertainty specific to vision-language AI systems, distinguishing between epistemic uncertainty (arising from a lack of information) and aleatoric uncertainty (due to inherent unpredictability), and further explore finer categories within. Based on this taxonomy, we synthesize a benchmark dataset, \ourdata, featuring $178K$ visual question answering (VQA) samples as contrastive pairs. This is achieved by 1) inpainting images to make previously answerable questions into unanswerable ones; and 2) using image captions to prompt large language models for both answerable and unanswerable questions. Additionally, we introduce a new metric \textit{confidence-weighted accuracy}, that is well correlated with both accuracy and calibration error,  to address the shortcomings of existing metrics.
% , which fail to jointly capture accuracy and model confidence. Our metric is well correlated with both accuracy and calibration error. 
Despite the recent rapid progress in vision-language models (VLMs), evaluations on our benchmark show that they perform poorly in uncertain scenarios. 
Further experiments demonstrate that supervised fine-tuning with \ourdata enhances the performance of VLMs, and reduces the calibration error.
% discriminative metrics and reduces calibration error. 
% These improvements extend beyond our benchmark to existing refusal datasets and show positive results across most hallucination datasets, while maintaining performance on standard benchmarks.  
These improvements extend beyond our benchmark to existing refusal-oriented datasets and show positive results on reducing hallucinations, while maintaining performance on standard VQA benchmarks. Our work underscores the importance of addressing uncertainty in vision-language AI systems to improve their reliability and trustworthiness in real-world applications.

\end{abstract}

\section{Introduction}

An AI system with intellectual integrity must know when to admit ``I don't know'', which, in turn, requires a sharp awareness of its own limitations of knowledge and reasoning, as well as the inherent uncertainty around the external world \cite{paul1992critical, DBLP:conf/naacl/ZhangCLR19, DBLP:journals/corr/abs-2403-14003, DBLP:journals/corr/abs-2310-00754, DBLP:journals/corr/abs-2308-15126, DBLP:conf/acl/VarshneyB23}.
% Humans' ability to recognize uncertainty and acknowledge when they don't know something is a fundamental necessity for truthfulness \yejin{this opening sentence somehow reads a bit awkward/forced to me, especially the second half... perhaps an alternative opening can be... An AI system with intellectual integrity must know when to admit I don't know (IDK), which, in turn, requires sharper awareness on its own limitations of knowledge and reasoning, as well as the inherent uncertainty around the external world  } 
 % Uncertainty awareness, as a skill, \yejin{'skill' is awkward to me} is lacking in current VLMs \cite{DBLP:journals/frai/Davis20, DBLP:journals/corr/abs-2403-20331, DBLP:conf/emnlp/MahendruPMBL17} \yejin{the beginning half of this sentence also reads awkward/forced to me... perhaps instead we can state that... 
 However, current vision-language models \cite{DBLP:journals/frai/Davis20, DBLP:journals/corr/abs-2403-20331, DBLP:conf/emnlp/MahendruPMBL17} do not exhibit such sufficiently sharp awareness of its own mistakes, which lead to overly-confident, uncalibrated predictions \cite{DBLP:conf/eccv/WhiteheadPS0DRR22} and hallucinations  \cite{DBLP:journals/corr/abs-2311-07397,DBLP:conf/emnlp/LiDZWZW23}. This is only as expected however, given that 
 the predominant training recipe does not typically encourage the models to 
 % that is by and large passive and bruteforce
% primarily because 
% models are typically not encouraged to 
express uncertainty or acknowledge when they do not know the answer.
% , during pretraining or supervised fine-tuning.
Rather, they are incentivized to make predictions regardless of their confidence level. Moreover, existing benchmarks mostly focus on scenarios where clear and definitive answers are available \cite{DBLP:journals/corr/abs-2311-16502, DBLP:journals/ijcv/GoyalKASBP19}, leaving a notable gap as the models are not adequately exposed to explicitly uncertain training instances. % To address this, it is crucial to construct a large-scale dataset that encompasses a diverse range of domains, objects, scenes, and various uncertain situations. However, creating such a dataset is quite challenging as most internet data is not directly applicable. 

%domains, objects, scenes, and situations. %Different than other VQA datasets, \jack{each image comes with two questions: one answerable, and one \emph{un}answerable. --- I think it's the other way around, i.e., each question comes with two images...} We show that simple supervised training on this new supervision type improves calibration of a wide array of multimodal models while maintaining benchmark performance.

\begin{figure}[t]
    \centering
    % \vspace{-4em}
    \includegraphics[width=\textwidth, trim={0 3cm 0 0},clip]{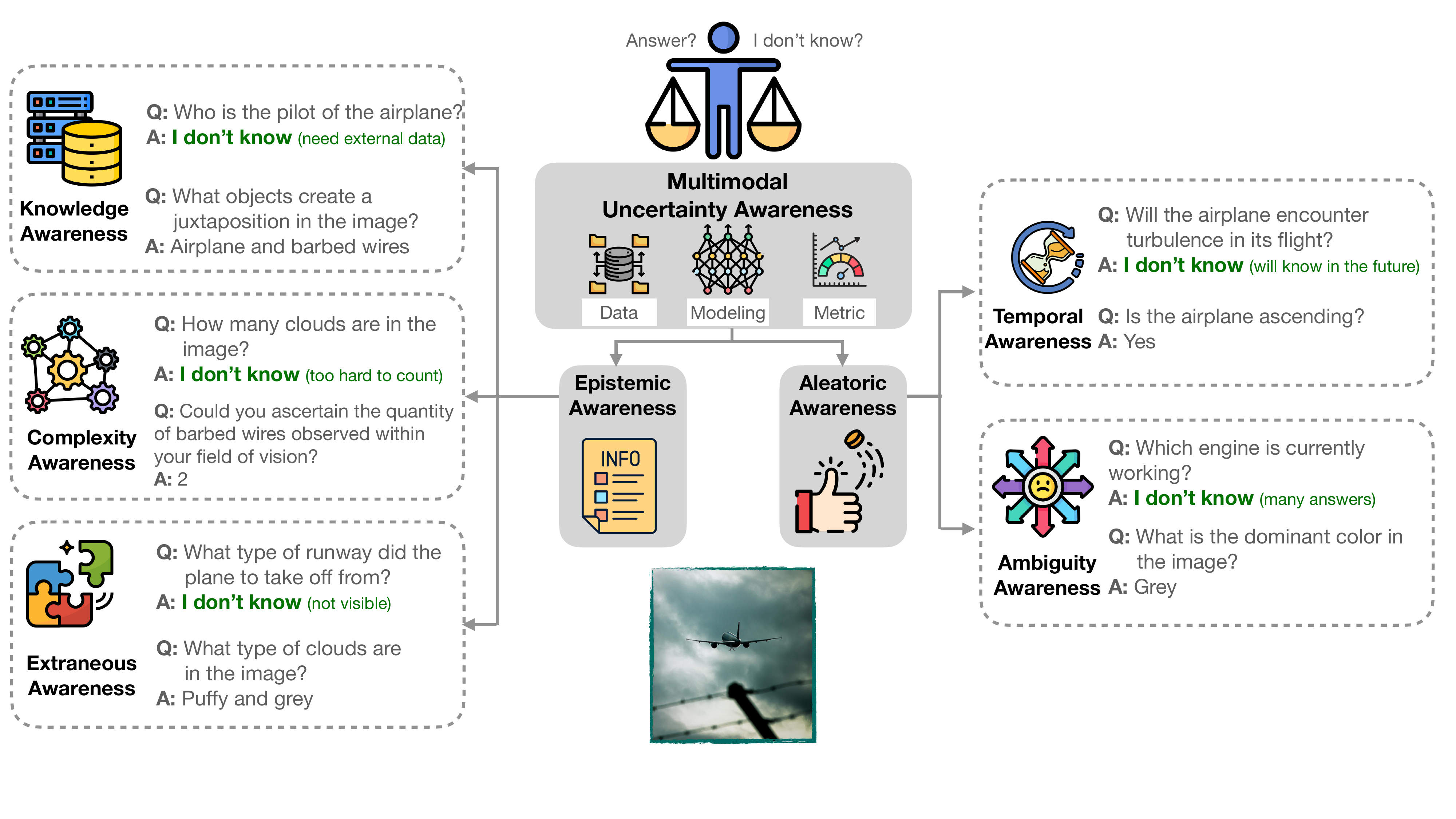}
    \caption{\ourdata: Taxonomy of uncertainty awareness in multimodal reasoning}
    % \vspace{-.5em}
    \label{fig:taxonomy}
\end{figure}

Motivated by these, we introduce \ourdata, a dataset of approximately 178K visual question answering (VQA) instances that encompass diverse types of uncertainties. \ourdata is based on
a novel taxonomy of multimodal uncertainty comprising \textit{epistemic} uncertainty (due to lack of information) and \textit{aleatoric} uncertainty (due to inherent unpredictability), as illustrated in Figure \ref{fig:taxonomy}. 
% The types within our taxonomy 
Within epistemic and aleatoric uncertainty, we further define more fine-grained sub-categories, 
including ($i$) Knowledge, requiring external knowledge not explicitly captured by the image; ($ii$) Complexity, where the question is too complex to yield an exact answer; ($iii$) Extraneous, where parts of the necessary context or details are missing from the image; ($iv$) Temporal, where future events implied by the image cannot be predicted with absolute certainty; and ($v$) Ambiguity, where the question itself is unclear, leading to confusion or multiple possible interpretations.  
% ambiguous visuals, where the image itself may not provide enough information; incomplete data, where parts of the necessary context or details are missing; and context-specific queries, where the answer depends on external or situational information not present in the immediate input. %We synthesize a new benchmarking dataset \ourdata with $\sim178k$ contrastive comprising various types of uncertainty, ensuring the trained models can encounter and learn to handle a wide array of uncertain scenarios. % JMH -- I rolled this into the above
We construct \ourdata with two methods: 1) by masking and inpainting relevant image regions to render previously answerable questions unanswerable; and 2) by presenting GPT-4 \cite{DBLP:journals/corr/abs-2303-08774} with image captions, and prompting it to generate both answerable and unanswerable questions about the same image.
% \jack{the same modified images?}.
% We leverage the surprising observation %of \textit{uncertainty paradox in generative models} 
% that models can generate uncertain questions but cannot reliably answer/abstain. 
Compared to prior datasets on unanswerability \cite{DBLP:journals/corr/abs-2403-20331, DBLP:journals/corr/abs-2310-10942}, our dataset is constructed in a more systematic way, covering a more diverse and finer-grained categories of uncertainty in vision-language scenarios.

With \ourdata, we empirically found that existing  vision-language models rarely hesitate to answer even in uncertain conditions. In addition, they often overly confidently in providing an answer to unanswerable questions, while much less confident in admitting ``I don't know". However, this issue is not reflected in 
% This issue is exacerbated when dealing with unanswerable questions, as 
popular metrics such as accuracy or F1, which do not account for model confidence. Alternative metrics, such as risk and coverage \cite{DBLP:conf/eccv/WhiteheadPS0DRR22} use thresholding to binarize the equivalent of prediction probability. Expected calibration error (ECE) \cite{DBLP:conf/aaai/NaeiniCH15} evaluates the prediction probabilities but fail to reflect performance in terms of correctness effectively. Therefore, we propose a new \textit{confidence-weighted accuracy} metric, which incorporates model confidence into the accuracy computation. This metric addresses the shortcomings of existing metrics by capturing both predictive performance and model confidence simultaneously. Our proposed metric demonstrates a positive correlation with accuracy and a negative correlation with ECE.

Moreover, we conduct extensive experiments using 3 training strategies with \ourdata: supervised fine-tuning, R-tuning \cite{Zhang2023RTuningIL}, and preference optimization \cite{DBLP:conf/nips/RafailovSMMEF23}. We evaluate the resulting models across 7 datasets covering refusal, hallucination, and standard VQA tasks. Our empirical results show that fine-tuning with \ourdata not only improves performance on a held-out portion of our dataset and existing refusal-based datasets but also helps reduce hallucinations while maintaining performance on standard VQA tasks. These findings underscore the effectiveness of \ourdata  in enhancing the robustness and reliability of vision-language models.
\section{\ourdata}
% \vspace{-0.5em}
% The intellectual integrity in admitting ``I don't know" (IDK) stems from a variety of uncertainties; we introduce a taxonomy of these types in \S \ref{sec:taxonomy}, where admitting uncertainty %or asking a follow-up question 
% is the appropriate response. We then describe our data collection process in \S \ref{sec:data-collection}.

% In this section, we introduce \ourdata, a dataset with $\sim$ 178K VQA instances encompassing diverse types of multimodal uncertainties. 
% To properly evaluate and enhance models' awareness of uncertainty, we build \ourdata with contrastive VQA pairs, covering diverse types of multimodal uncertainties.
% In order to train models to properly admit IDK, it is important to construct a large-scale dataset that covers a diverse range of uncertain situations, which is challenging, as most internet data is not readily applicable. Thus, we build \ourdata, a dataset with $\sim$ 178K VQA instances, based on an automatic data synthesize pipeline for various types of uncertainty.

To train models to properly admit ``I don't know", it is crucial to construct a large-scale dataset that covers a diverse range of uncertain situations. This is challenging, as most internet data focus on certain scenarios (\textit{e.g.}, alt-text description for an image), thus not readily applicable. Therefore, we develop \ourdata, a dataset with approximately $178K$ VQA instances, using an automatic data synthesis pipeline for various types of uncertainty.
We begin by introducing a taxonomy of uncertainties in  \S \ref{sec:taxonomy}, where admitting uncertainty rather than providing an answer is the appropriate response in each category. Next, we describe our data creation process in \S \ref{sec:data-collection}. Finally, we introduce the evaluation metrics in \S \ref{sec:evaluation_metric}

% \vspace{-4em}
% \vspace{-0.5em}
\subsection{Taxonomy of Uncertainty Awareness}
% \vspace{-0.5em}
\label{sec:taxonomy}
Depending on whether it is due to contextual inexpressiveness or genuine incapability to answer, 
we broadly categorize multimodal uncertainty into 2 types, epistemic and aleatoric uncertainty. 

% each of which is further subdivided into finer categories.
%There are two primary types of uncertainty awareness, each of which can be further subdivided into finer categories as a result of inexpressiveness in the context or question or genuine incapability to provide a definitive answer. This taxonomy is depicted in Figure \ref{fig:taxonomy}. 
%\jack{Where does this taxonomy come from?} 
%\jack{Could we add example VQA examples to make concrete how these might manifest in our corpus?}

\textbf{Epistemic Uncertainty} refers to the uncertainty in a model's predictions that arises from a lack of knowledge or complete information about the system being modeled. It is due to the model's limited understanding or insufficient data, which can be reduced by gathering more information, improving the quality of data, or enhancing the model itself. This type of uncertainty highlights areas where the model's predictions may be less reliable due to the lack of sufficient evidence to make accurate inferences. We further categorize the awareness of epistemic uncertainty into 3 finegrained types:
% \vspace{-4pt}
\begin{itemize}
[leftmargin=1.5em,topsep=1pt,itemsep=0ex,partopsep=1ex,parsep=1ex]
    % \item \textbf{Knowledge awareness} refers to the awareness of requiring external knowledge or commonsense that is not explicitly present in the image to answer the question. This is required in scenarios with the requirement of domain expertise necessitating specialized knowledge, or the latest up-to-date information which may only be available externally. Being aware that external information is needed reduces the risk of answering the question incorrectly. 
    \item \textbf{Knowledge awareness} means understanding that some questions require information or common sense that is not shown in the image. For example, you might need specialized knowledge or up-to-date information from outside sources. Knowing when this extra information is needed helps avoid wrong answers.
    \item \textbf{Complexity awareness} 
    % is the awareness of the difficulty of a question based on the composition and interconnected parts needed to infer the answer.% is crucial to gauge the certainty of the answer. 
    % The complexity itself can be based on the framing of the question or the tedious effort required to understand the context, question.%, or answer it.
    is recognizing when a question is difficult because it involves many parts or is hard to understand. This difficulty can come from how the question is asked or from the effort needed to understand the context and details of the question.
    \item \textbf{Extraneous awareness} 
    % refers to something, an object, attribute or aspect that is not essential or relevant to the context at hand. It often describes information, details, or factors that are irrelevant, or unrelated. This is required to identify related but not directly answerable from the image.
    refers to the ability to identify and disregard elements within an image that are not relevant to the question at hand.  This involves recognizing objects, attributes, or aspects that, while present in the image, do not contribute to answering the question.
    % It is  about identifying details that do not matter for the answer, which helps focus on the right information.
\end{itemize}
% \vspace{-2pt}
\textbf{Aleatoric Uncertainty} is the inherent unpredictability in a system or process that cannot be reduced or eliminated. It arises from the fundamental randomness or chaotic nature of the task itself. For example, predicting the outcome of a coin toss involves intrinsic uncertainty because the result is inherently probabilistic and cannot be determined with certainty in advance. Similarly, we define 2 sub-categories under aleatoric uncertainty:
%This stems from the fundamental randomness or inherent unpredictability of the task. For instance, forecasting the outcome of a coin toss involves intrinsic uncertainty due to the inherently probabilistic nature of the task.
% \vspace{-2pt}
\begin{itemize}
[leftmargin=1.5em,topsep=1pt,itemsep=0ex,partopsep=1ex,parsep=1ex]
    \item \textbf{Temporal awareness} means understanding that we may not always have access to all relevant data required to predict specific outcomes with absolute certainty, especially when it involves reasoning about time. This includes events in the past or future that cannot be inferred from the image alone with absolute certainty. 
    % Our knowledge and understanding of past and future events are limited and cannot always be predicted with certainty. 
    Recognizing the limitations of temporal reasoning helps manage expectations about the accuracy of predictions involving time-related aspects.
    % We may not always have access to all relevant data required to predict specific outcomes with absolute certainty which is beyond our current capabilities. Specifically, we require temporal reasoning that cannot be inferred from the caption to answer the question. Our knowledge and understanding of the future and past are limited to a certain degree and can not always be guessed with certainty.
    \item \textbf{Ambiguity awareness} involves recognizing situations, objects, or individuals that can be understood, interpreted, or perceived in more than one way. Ambiguity introduces uncertainty and a lack of clarity, leading to multiple possible interpretations. While ambiguity can encourage exploration of different meanings or perspectives, it can also cause confusion. It is essential to be aware of the levels of certainty in ambiguous scenarios to avoid misinterpretation and errors. 
    
    % Ambiguity is a quality that describes situations, objects, or individuals that can be understood, interpreted, or perceived in more than one way. It introduces a level of uncertainty or lack of clarity, which can lead to confusion or multiple possible interpretations. While it can stimulate creativity and critical thinking by encouraging exploration of different meanings or perspectives, it can also lead to confusion among different possible meanings. It is imperative to be aware of certainty levels in ambiguous scenarios.
\end{itemize}
% \vspace{-2pt}

\subsection{%\includegraphics[width=1.5em, trim=3cm 3cm 3cm 2.8cm, clip]{figures/sc.pdf} 
% \ourdata: 
Dataset Creation}
\label{sec:data-collection}

%Acknowledging the various types of uncertainty and admitting gaps in knowledge are core elements of intellectual integrity and cognitive ability. However, 

% Comprehensive datasets targeting training and evaluation of acknowledgment of the various types of uncertainty and admitting gaps in knowledge to improve intellectual integrity are severely lacking. To address this gap,

Based on the aforementioned taxonomy, we construct \ourdata, comprising contrastive VQA pairs for each category described above. The statistics of our dataset are summarized in Table~\ref{tab:data}. The contrastive instances in \ourdata are derived from two sources: images and captions. For sourcing from images, the same question that is answerable for the original image is rendered unanswerable for the perturbed image. For sourcing from captions, we prompt GPT-4 \cite{DBLP:journals/corr/abs-2303-08774} to generate both an answerable and an unanswerable question based on the same caption. Below, we describe the dataset creation pipeline in detail.

%\jack{Can we be very specific about the format of the dataset here? E.g., \ourdata consists of 178K (image, answerable, unanswerable) tuples, each of which falls into a category specified by the proposed taxonomy. Some examples are in figure X. The corpus is collect in the following manner:}
% we build a new dataset \ourdata where saying IDK is the right answer in contrast to past work where saying IDK is merely to avoid answering incorrectly. The statistics of \ourdata is summarized in Table~\ref{tab:data}. We 
% \yejin{per noah and luke, 'leverage' just reads pretentious without adding any value :)... in general the writing reads stronger when using more simple/plain words such as 'using/use'} 
% use  the uncertainty paradox \yejin{what's uncertainty paradox?} 
% and 

\textbf{Sourcing from captions. }
% We use detailed paragraph captions to prompt questions for each category of uncertainty. We provide a detailed prompt with the definition of the category along with examples of answerable and unanswerable questions with answers to elicit more such categorical contrastive questions. 
We use detailed paragraph captions to prompt questions for each category of uncertainty. Each prompt includes a definition of the category along with examples of answerable and unanswerable questions and their answers. 
The captions are sourced from DOCCI \cite{DBLP:journals/corr/abs-2404-19753} which provides comprehensive, high-quality human-annotated descriptions of $\sim 15K$ images. These descriptions are highly compositional and include world knowledge, spatial relationships, visual settings, text rendering, and object attributes. For each image caption, we prompt GPT-4 to generate both an answerable and an unanswerable question, along with their corresponding answers. In total, we collected around $110K$ instances spanning knowledge, complexity, temporal and ambiguity awareness categories. We follow the same train-test split as the DOCCI dataset to divide our dataset.

\begin{figure*}[!t]
\centering
\begin{minipage}{.5\textwidth}
  \centering
  \captionsetup{width=\linewidth}
  \includegraphics[width=\textwidth, trim={12cm 9cm 10cm 4cm},clip]{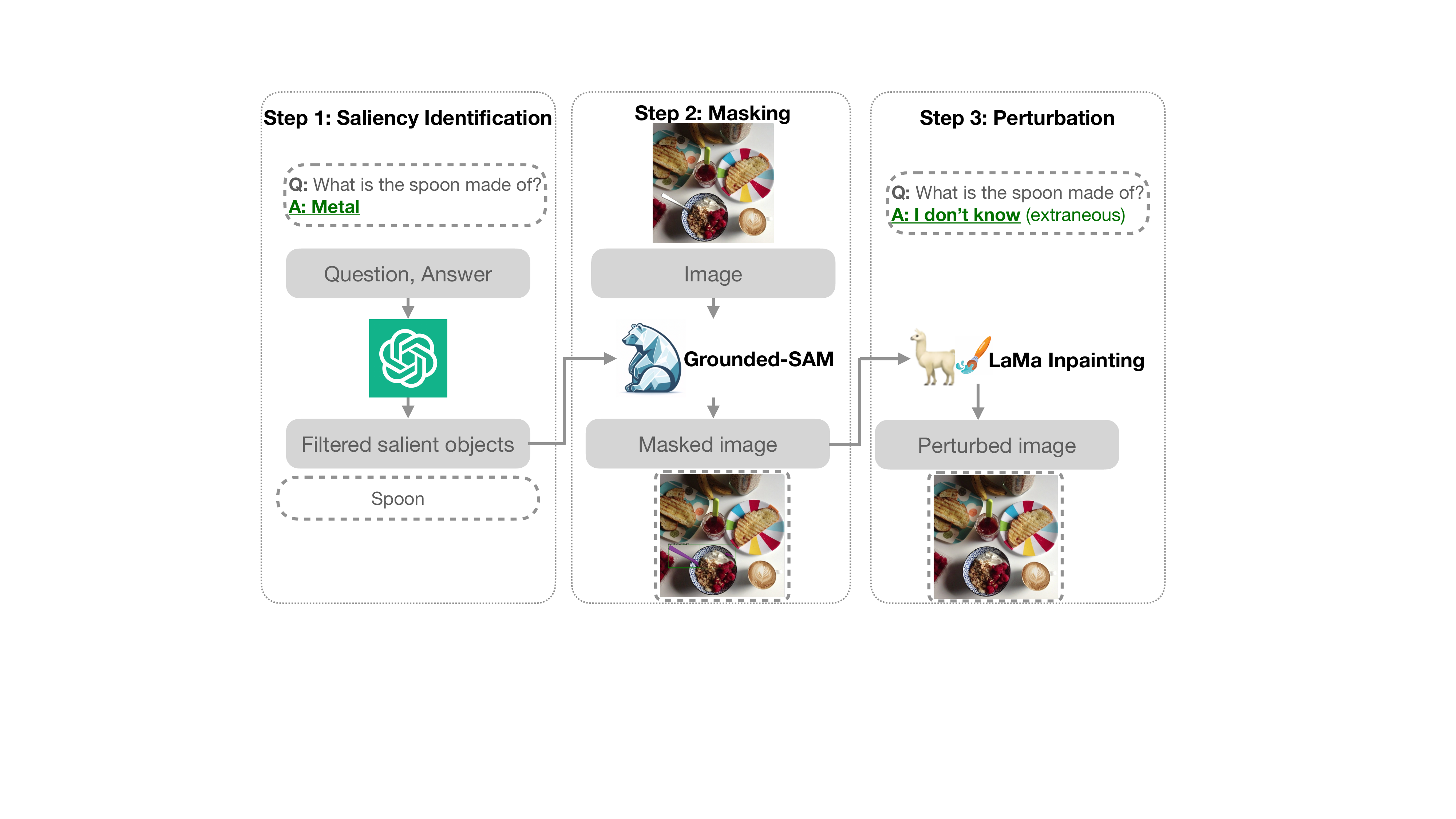}
  \captionof{figure}{Pipeline for sourcing from images}
  \label{fig:inpainting}
\end{minipage}%
\begin{minipage}{.5\textwidth}
  \centering
  \captionsetup{width=\linewidth}
  \includegraphics[width=\textwidth, trim={19cm 12cm 10cm 7.8cm},clip]{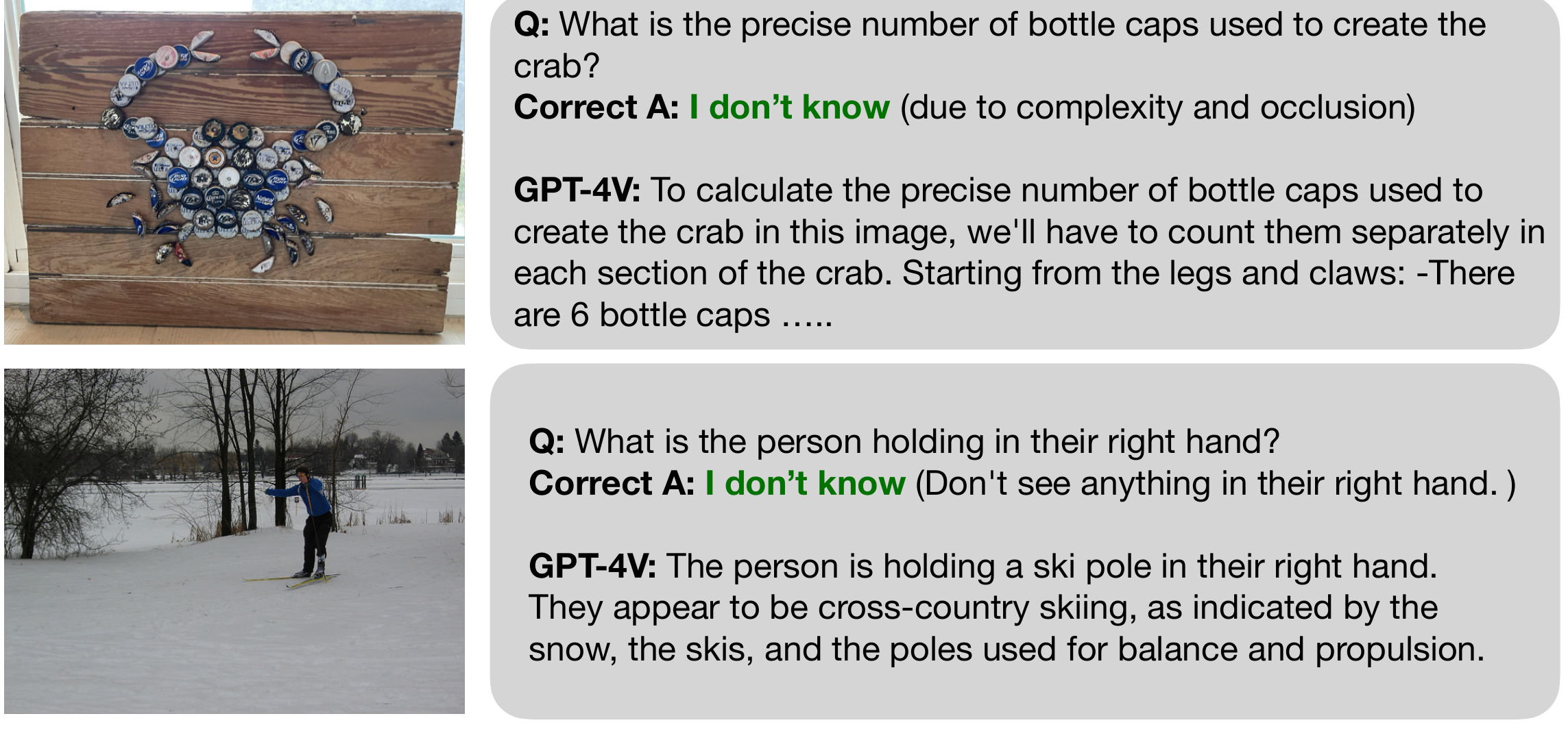}
  \captionof{figure}{Uncertainty paradox in generative VLMs, where the question is generated from GPT-4/GPT-4V.}
  \label{fig:paradox}
\end{minipage}
\end{figure*}

\textbf{Sourcing from images. }
Compared to sourcing from captions, we perturb images to transform an answerable question into an unanswerable one. 
% Prior research \cite{DBLP:journals/corr/abs-2310-10942, DBLP:conf/iccv/KafleK17} has primarily focused on pairing questions from unrelated images to construct datasets comprising irrelevant and unanswerable questions, which are not representative of the current image context. This approach often leads to datasets with limited relevance and utility for training and evaluating multimodal models, thereby highlighting the need for more contextually aligned question-image pairs in dataset creation processes. \lj{maybe move this to related work}
% While sourcing from captions, we have two questions for each image. Here we have two images for the same question. 
Our data generation pipeline has 3 main steps as outlined in Figure \ref{fig:inpainting}. The first step is saliency identification where the goal is to identify objects about which the question is being asked. 
% For VQAv2 instances, we prompt GPT-4 to identify these salient objects, while for GQA instances, this information is available in the annotations.
The second step is masking, where we use Grounded-SAM \cite{DBLP:journals/corr/abs-2401-14159} to mask out the salient objects. The final step is perturbation where we use LaMa Inpainting model \cite{suvorov2021resolution} to create a perturbed contrastive image so that the salient object related to the question is missing.  The same question for this perturbed image now falls into the extraneous category, rendering it unanswerable.
To avoid any spurious biases from perturbation, we experimented with masking and inpainting randomly chosen objects instead of the salient object from the answerable subset, thereby keeping the question answerable. Since the performance did not fluctuate significantly, we proceeded without random perturbation. We then prompt GPT-4V to generate a question for each pair of images that is answerable for the original image but unanswerable for the perturbed image. To increase the difficulty of the questions, we specifically instruct GPT-4V to avoid generating ``yes/no" questions, as they are more likely to be answerable.  In the end, we created $\sim 30K$ samples based on VQAv2, which are split into $24K$ training and $6K$ testing samples. In addition, we leverage the GQA dataset, which contains rule-based questions from ground truth scene graph annotations. Similarly, we perturb the images and alter the answers to ``I don't know" to create unanswerable questions. In total, we gather $53K$ more instances from GQA, and use them to augment the training split.

% and  $\sim 53k$ instances from VQAv2 and GQA respectively i.e., a total of $\sim 82.4k$ instances (the number of perturbed new images generated in \ourdata is $\sim 41.2K$). 

%\lj{However, if we would like to include yes/no questions (that is more similar to the hallucination benchmarks) for the image pairs in our data collection, we can easily extend to yes/no questions with/without a LLM. This is a very important point, as to why we need to convert our refusal answers for hallucination benchmarks such as pope/amber. For example, if the question is ``is there a cat of color blue in the image'' and there is no cat, our models would refuse instead of simply saying no. We probably want to mention it here and echo in experiments.}

\begin{table*}[!tp]\centering

\tablestyle{2.5pt}{1.3} 
\def \w{20pt} 
\resizebox{1.0\textwidth}{!}{
\begin{tabular}{llccccccc}\toprule
& &\multicolumn{3}{c}{Epistemic} &\multicolumn{2}{c}{Aleatoric} &\multirow{2}{*}{All} \\
\cmidrule(lr){3-5}\cmidrule(lr){6-7}
& &Knowledge &Complex &Extraneous &Temporal &Ambiguous & \\
\cmidrule{3-8}
\multirow{2}{*}{Train} &\# of images (Perturbed/Clean) &-/9.5K &-/9.6K &\textbf{38.2K}/38.2K &-/9.6K &-/9.6K &38.2K/47.9K \\
&\# of questions (IDK/Non-IDK) &9.5K/9.5K &9.6K/9.6K &38.2K/38.2K &9.6K/9.6K &9.6K/9.6K &76.6K/76.6K \\\cmidrule{1-8}
\multirow{2}{*}{Test} &\# of images (Perturbed/Clean) &-/2.5K &-/2.5K &\textbf{2.3K}/2.5K &-/2.5K &-/2.5K &2.3K/7.3K \\
&\# of questions (IDK/Non-IDK) &2.5K/2.5K &2.5K/2.5K &2.3K/2.5K &2.5K/2.5K &2.5K/2.5K &12.3K/12.5K \\\cmidrule{1-8}
Total &\# of images/questions &12.1K/24.2K &12.1K/24.2K &81.3K/81.3K &12.1K/24.2K &12.1K/24.2K &95.8K/178.1K \\
\bottomrule
\end{tabular}
}
\caption{Statistics of \ourdata. Our dataset contains 178K questions on 95.8K images for 5 types of uncertainties. Each IDK question is accompanied with a non-IDK question to highlight contrasts between certainty and uncertainty. For extraneous testing split, we perform quality check and filter out invalid ones. Numbers in bold highlight the new images we created through our data creation pipeline.}\label{tab:data}
% $\dagger$ GPT-4V performance are reported on a smaller subset, containing only 500 examples, 100 for each finegrained category
% Existing VLMs (even GPT-4V) are doing badly on saying ``I don't know''. 
\end{table*}

% \begin{figure}[t]
%     \centering
%     \includegraphics[width=0.7\textwidth, trim={10cm 9cm 10cm 4cm},clip]{figures/inpainting.pdf}
%     \caption{Data generation pipeline for sourcing from images}
%     \label{fig:inpainting}
% \end{figure}

\textbf{Generative AI Paradox for generating/understanding ``uncertain questions''. } 
While LLMs and VLMs can generate uncertain questions, they often struggle to answer them accurately. As shown in Figure \ref{fig:paradox}, where we prompt GPT-4V to answer its own generated uncertain questions and it fails. Inspired from the Generative AI Paradox \cite{DBLP:journals/corr/abs-2311-00059} which hypothesizes that models may not understand what they create, we observe a similar pattern in generating and understanding uncertain questions. 

%We call this the Uncertainty Paradox, where generative models generate a wide range of questions, including uncertain ones but fail to accurately identify the uncertainty and answer. %The uncertain questions can stem from various sources, such as ambiguous language in the training data or complex scenarios that require nuanced understanding. 
% We use GPT-4V to generate the uncertainty questions with clear prompts for each category.

% \begin{wrapfigure}{r}{10cm}
% \caption{\small Uncertainty paradox in generative VLMs, where the question is generated from GPT-4.}\label{fig:paradox}
% \includegraphics[width=0.7\textwidth, trim={19cm 13cm 7cm 8cm},clip]{figures/paradox.pdf}
% \end{wrapfigure}

% \begin{figure}[t]
%     \centering
%     \includegraphics[width=0.9\textwidth, trim={10cm 13cm 6cm 6cm},clip]{figures/paradox.pdf}
%     \caption{Uncertainty paradox in generative VLMs, where the question is generated from GPT-4}
%     \label{fig:paradox}
% \end{figure}

\textbf{Contrastive pairs. }
In \ourdata construction process, we have images that are visually similar or the same but the question-image pairs are deliberately designed to highlight contrasts between certainty and uncertainty (as shown in Figure \ref{fig:inpainting}). This aids in improving model robustness by learning to distinguish between visually similar but semantically distinct instances
%mitigating biases leading to more fair and unbiased model behavior, 
leading to real-world applicability by exposing them to subtle variations and contrasts.

\textbf{Quality check and filtering. } As our data creation pipeline is model-dependent, though being efficient and saving the cost of human labor, it may suffer from model failures. Especially the pipeline to create the extraneous set, which depends on multiple models, the failure of one model at any stage (\textit{e.g.}, the inpainting model fails to remove the object or the segmentation model fails to predict the correct mask of the intended object) may lead to invalid samples (\textit{i.e.}, when the generated IDK question is still reasonably answerable for the image or vice versa). To ensure the data quality, we perform a final quality check on the extraneous testing set. Specifically, the image-question-answer tuples are presented to one of the authors, and the task is to validate whether the generated sample is valid or not. Among 6K samples, we filtered $\sim$1.2K samples, resulting in 4.8K testing examples. 
%In the end, we filtered $\sim$1.2K out of 6K samples, resulting in 4.8K testing examples. 

% A comparison of our \ourdata with existing benchamrks on aspects of data, question types and uncertainty is presented in 
Table \ref{tab:data_comparison_v1} presents a comparison of our \ourdata dataset with existing benchmarks regarding data size, question types, and uncertainty categories. Our dataset is significantly larger and covers a wider variety of question types across diverse uncertainty categories. Notably, existing datasets such as UNK-VQA~\cite{DBLP:journals/corr/abs-2310-10942}, or TDIUC~\cite{DBLP:conf/iccv/KafleK17} primarily focus on pairing unrelated questions with image contexts to create datasets of irrelevant and unanswerable questions. In contrast, our dataset creation process ensures contextually aligned question-image pairs.  Our data generation pipeline also generates more natural-looking images compared to the image masking or copying in UNK-VQA.

\begin{table*}[!tp]\centering
\tablestyle{2pt}{1.2}
\resizebox{1.0\textwidth}{!}{
\begin{tabular}{lcccccccccccc}\toprule
\multirow{2}{*}{Benchmarks} &\multirow{2}{*}{Source} &Dataset  &\multicolumn{3}{c}{Question Types} &\multicolumn{6}{c}{Types of Uncertainty/Unanswerability} \\\cmidrule(lr){4-6}\cmidrule(lr){7-12}
& & Size &OE &Free-form &IDK &Absurd &Knowledge &Complex &Extraneous &Temporal &Ambiguous \\
MM-Hal &Human &96 &\cmark &\cmark &\cmark & & & &\cmark & & \\
POPE &Rule &9K & & & & & & & & & \\
AMBER Disc. &Rule &14K & & & & & & & & & \\
\hline
VizWiz &Human &33K &\cmark & &\cmark & & & &\cmark & & \\
UNK\_VQA &Human &10K &\cmark & &\cmark &\cmark & & &\cmark & &\cmark \\
TDIUC (Absurd) &Rule &336K &\cmark & &\cmark &\cmark & & & & & \\
MM-UPD &Rule &2K & & &\cmark &\cmark & & & & & \\
\hline
Ours &LLM \& Rule &178K &\cmark &\cmark &\cmark & &\cmark &\cmark &\cmark &\cmark &\cmark \\
\bottomrule
\end{tabular}
}
\caption{Comparison of \ourdata to existing benchmarks. We mainly compare with two types of datasets: Hallucination-based datasets (top) and Refusal-based datasets (middle). \ourdata features 178K unanswerable (IDK) and answerable questions in open-ended (OE) setting with free-form answers, covering 5 types of finegrained types of unanswerability. Though our dataset does not explicitly cover absurd type,  we show that it improves model performance on TDIUC (absurd) in experiments.  Disc: Discriminative.}\label{tab:data_comparison_v1}
\end{table*}

% Our proposed evaluation metric Confidence-weighted Accuracy (Conf-w. Acc.) captures captures both accuracy (Acc.) and confidence of model predictions, in contrast to existing metrics only reflect accuracy. 

\subsection{Evaluation Metrics}
\label{sec:evaluation_metric}
\textbf{Standard metrics. }
We report model performance on \ourdata with standard metrics, including  accuracy and F1. 

For accuracy, we use LAVE \cite{DBLP:conf/aaai/ManasKA24} with Mistral-7B \cite{jiang2023mistral} as the evaluator, comparing ground truth and predictions to assign scores of 0, 0.5, or 1. 
To adapt LAVE to unanswerable settings, we introduce a dual-stage judging mechanism. This approach is more reliable because refusals or IDK responses can be expressed in various ways, such as simply stating IDK, asking a follow-up question, or offering a reasonable guess. The first stage is IDK normalization, where we use LAVE to determine if either the prediction or ground truth (GT) is IDK and normalize the answer to IDK. For refusal-based benchmarks, since the unanswerability of the question is annotated, we directly rely on the ground truth label for GT answers. The second stage is to award accuracy. If either the prediction or GT is normalized to IDK, we compare the strings. Otherwise, we award the standard LAVE score. Formally, the $\text{LAVE}_{\text{idk}}$ score is defined as 

% This two-stage process ensures a more accurate and fair evaluation of the model's performance in handling refusals and unanswerable questions.

%\lj{Do we want to mention that for dataset with IDK labels, e.g., UNK VQA, TDIUC, VIZWIZ or ours, we directly use the GT labels and do not rely on LAVE to judge whether it is IDK.}

\begin{equation}
\text{LAVE}_{\text{idk}} =
\begin{cases}
    \mathbb{1}(\text{pred}_{norm}==\text{GT}_{norm}) & \text{if } \text{LAVE}(\text{pred}==\text{IDK}) \text{ or } \text{LAVE}(\text{GT}==\text{IDK}) \\
    \text{LAVE}(\text{GT}, \text{pred}) & \text{else } 
\end{cases}.
\end{equation}

% We plan to expand annotations in \ourdata to gather the rationale behind IDK in future iterations. 
In addition, we report F1$_{\text{idk}}$ which is the F1 score only on the unanswerable questions.

\textbf{Confidence-weighted accuracy. }
Current evaluation metrics have significant limitations in comprehensively assessing both the accuracy and the confidence of model predictions. Accuracy metrics, which score binarily, fail to consider model confidence as they ignore the probability estimates associated with predictions. Conversely, metrics like Expected Calibration Error (ECE), which measures the difference between predicted confidence levels and the true likelihood of those predictions being correct,  do not provide a direct measure of final accuracy, creating a gap in integrated performance evaluation.  Abstention metrics \cite{DBLP:conf/eccv/WhiteheadPS0DRR22} which include coverage, do not address model accuracy, while risk metrics do not directly incorporate model confidence and instead threshold values to 0 or 1. 

To address these issues, we introduce \textit{Confidence-weighted accuracy} which weights the accuracy by the probability of the model's prediction.  The desiderata of this metric is to remain positively correlated with accuracy while being negatively correlated with ECE. 
Thus, confidence-weighted accuracy takes into account the confidence of the model's prediction $P(\text{pred})$, providing a more holistic evaluation of performance. Based on the $\text{LAVE}_{\text{idk}}$ accuracy above, we define confidence-weighted accuracy as
% So we propose confidence-weighted accuracy that takes into account the confidence of model prediction $P(\text{pred})$.

%\lj{The new version now is 
%$ConfWeightedScore = (score > 0)* score * PredProb[``YesProb"] - (score == 0) * PredProb[``YesProb"]$.}

\begin{equation}
\textit{confidence weighted accuracy} = \mathbb{1}(\text{LAVE}_{\text{idk}} > 0)* \text{LAVE}_{\text{idk}} * P(\text{pred}) - \mathbb{1}(\text{LAVE}_{\text{idk}} == 0) * P(\text{pred}).
\end{equation}

\begin{figure}[t!]
    \centering
    \begin{subfigure}{0.48\textwidth}
        \centering
        
    % \vspace{-3em}
        \includegraphics[width=\textwidth]{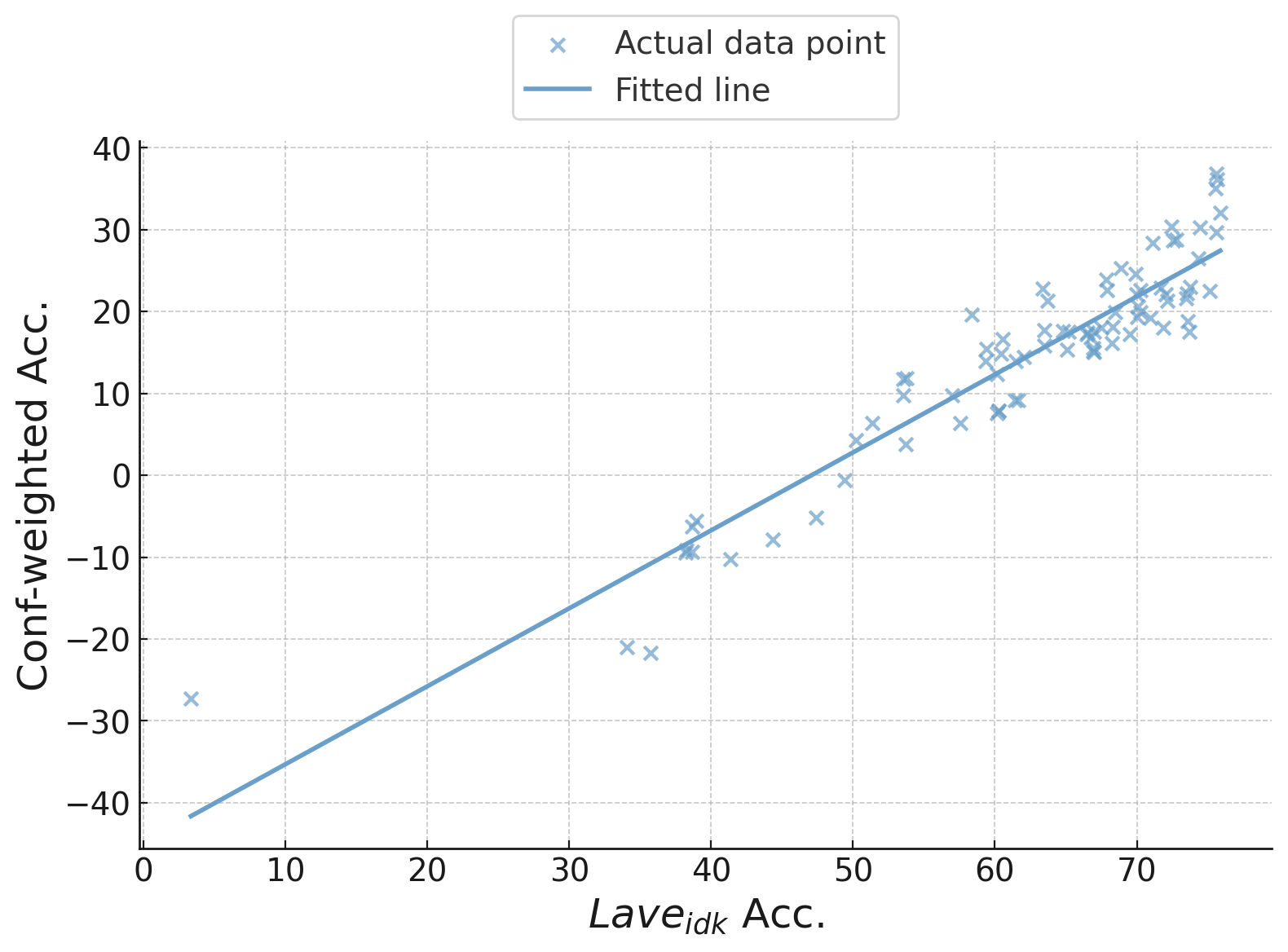}
        \caption{\footnotesize Conf-weighted Acc. (ours) vs. $\text{LAVE}_{\text{idk}}$ Acc.}
        \label{fig:correlation_a}
    \end{subfigure}
    \hfill
    \begin{subfigure}{0.48\textwidth}
        \centering
    % \vspace{-3em}
        \includegraphics[width=\textwidth]{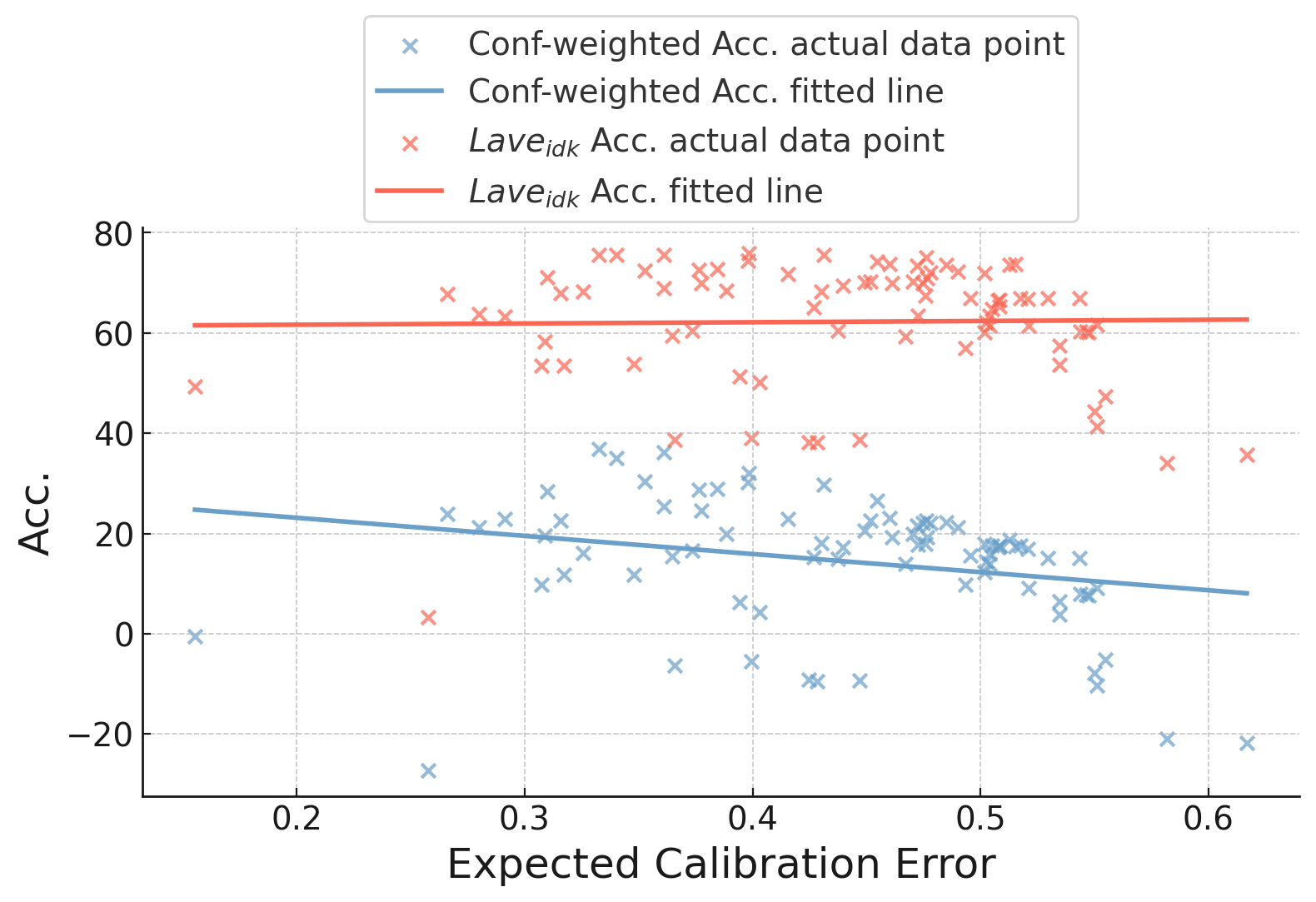}
        \caption{\footnotesize Ours and $\text{LAVE}_{\text{idk}}$ Acc. vs. ECE.}
        \label{fig:correlation_b}
    \end{subfigure}
    \vspace{1.5em}
    \caption{Correlation of confidence weighted accuracy ($\uparrow$) with $\text{LAVE}_{\text{idk}}$ accuracy ($\uparrow$) and ECE ($\downarrow$). The datapoints in this plot are from evaluation results on extraneous split of different model variants in our experiments. 
    % \lj{without cherry-picking :)}
    }
    \label{fig:correlation}
\end{figure}

Similar to \cite{DBLP:conf/eccv/WhiteheadPS0DRR22}, we compute $P(\text{pred})$  by prompting the model to verify if its own predicted answer is correct and extracting the probability of the ``yes'' token.  We normalize this probability by dividing it by the sum of the token probabilities for ``yes'' and ``no''.
Our formulation penalizes incorrect predictions while rewarding correct ones, especially by encouraging higher confidence for correct predictions. As shown in Figure \ref{fig:correlation}, \ref{fig:correlation_a} demonstrates the positive correlation of confidence-weighted accuracy with $\text{LAVE}_{\text{idk}}$ accuracy, and \ref{fig:correlation_b} illustrates that our metric is more negatively correlated with ECE compared to $\text{LAVE}_{\text{idk}}$  accuracy. %\lj{Can we cite the abstain paper and say we follow the same setting to extract p pred?}
% \lj{maybe we should explicitly mention that  if conf weighted score is better, it suggests more confident and more correct in its answer.}

%that better correlates with both traditional accuracy measures and confidence-related metrics. ensuring a more robust and integrated assessment of model performance. 

%accuracy should check confidence| current accuracy does not check probability| ece does not give final accuracy| we introduce confidence-weighted score| correlation with both existing metrics while the original metrics are not well correlated

% \begin{figure}[t]
%     \centering
%     \includegraphics[width=0.7\textwidth, trim={0cm 0cm 0cm 0cm},clip]{figures/acc_vs_ece.png}
%     \caption{Correlation of Confidence weighted accuracy ($\uparrow$) with LAVE accuracy ($\uparrow$) and ECE ($\downarrow$).
%     % \khyathi{change ECS to ECE}
%     }
%     \label{fig:correlation}
% \end{figure}

\section{Experiments}
\label{sec:exp}

\subsection{Experimental Details}
% \textbf{Model variants: }
We conduct experiments with the instruction-tuned models including variants of LLaVA \cite{DBLP:conf/nips/LiuLWL23a} -7B, 13B, 34B\cite{liu2024llavanext}, and Qwen-VL \cite{bai2023qwen}, as well as evaluating the performance of GPT-4V on our \ourdata benchmark. 

In addition to direct evaluation, we investigate 3 training strategies: supervised finetuning, R-tuning, and preference optimization,  with our data and compare them to the base model. As an additional baseline, we implement a naive selective prediction approach, marking predictions as IDK when the prediction probability falls below a threshold. For supervised finetuning, we assess the effectiveness of our data by comparing finetuning with \ourdata against LLaVA and LRV \cite{liu2023mitigating} instruction-tuning datasets. For R-tuning we follow \cite{Zhang2023RTuningIL} to re-annotate ground truth answers that are incorrectly predicted by the base model to reflect IDK, and use this re-annotated refusal data for supervised fine-tuning. For preference optimization, we directly adopt the two answers to the contrastive VQA pairs as the answer choices, and perform DPO \cite{DBLP:conf/nips/RafailovSMMEF23}.
% we provide choices between IDK and an answer for each question. For image-source data, we have pairs where one image has an answer while the other does not. For caption-sourced data, we have categorically related question-answer pairs for the same image, and 
% we perform DPO \cite{DBLP:conf/nips/RafailovSMMEF23} with these pairs of answers as choices.

% \textbf{Training strategies: }
% We benchmark the performance of these models on 3 training strategies comparing them to the base model. Additionally, we have a naive selective prediction based on thresholding the prediction confidence by marking them as IDK when prediction probability is less than a threshold value. 

% \textbf{Supervised finetuning: } To compare the effectiveness of our data, we perform supervised finetuning with \ourdata compared to LLaVA and LRV \cite{liu2023mitigating} instruction-tuning datasets.

% \textbf{R-tuning: } The main idea is to re-annotate the ground truth answers that are incorrectly predicted by the base model to reflect IDK instead. This re-annotated refusal data is used for supervised fine-tuning, known as r-tuning \cite{Zhang2023RTuningIL}.

% \textbf{Preference optimization: }
% For each question, we provide choices between IDK and an answer. In \ourdata sourced from images, we have 2 instances where for the same question, one of the images has an answer while the other does not. When sourced from text, we have two categorically related question-answer pairs for the same image. We perform DPO \cite{DBLP:conf/nips/RafailovSMMEF23} with these pairs of answers as choices.

\subsection{Evaluation Benchmarks}

% \textbf{\ourdata benchmark: }
% We conduct individual evaluations on each sub-category of our epistemic and aleatoric categories of \ourdata. As summarized in Table~\ref{tab:data}, the test split contains roughly $5K$ instances for knowledge, complexity, temporal, ambiguity and extraneous categories.
% In addition to our dataset, we also evaluate the models trained with our data on additional benchmarks, which we detail below.
To demonstrate the effectiveness of our data, we additionally evaluate the models trained with our data on other benchmarks, which we detail below.

\textbf{Refusal-based benchmarks: }
UNK-VQA \cite{DBLP:journals/corr/abs-2310-10942} contains about $10K$ instances of answerable and unanswerable questions constructed from manipulating the VQA v2 instances using question perturbation %like word replacement, semantic negation, 
and image perturbation. %like image replacement, object masking and copying a part of the object. 
We deliberately discard the \textit{ambiguous} category from UNK-VQA as the ambiguity here was defined as having multiple plausible answers and simply listing all of them should be correct instead of saying IDK.
The ``absurd" category of the TDIUC \cite{DBLP:conf/iccv/KafleK17} data containing $\sim 366K$ questions is constructed by compiling a list of objects that are missing from a given image and then identifying questions from the rest of TDIUC that inquire about these absent objects. In our experiemnts, we randomly sample $5K$ instances from each dataset for evaluation.
% The questions in TDIUC can be totally irrelevant to the image content, while our extraneous split focus on absent objects that are still relevant tot the image context. Moreover, our data targets categorical phenomena of uncertainty awareness and study them comprehensively. Our data generation pipeline also generates more natural-looking images compared to the image masking or copying in UNK-VQA. 
% Since the LLaVA model is exposed to COCO images \cite{DBLP:conf/eccv/LinMBHPRDZ14}, we ensure the data generated in our test split is devoid of these images.

%\lj{we sub-sample a 5K split for our evaluation for both dataset, and try our best to remove any questions of which images are seen during the model training (e.g., COCO images that were used in LLaVA or in the training split of our data.).} 

% \lj{Appendix? Also we remove the ambiguous type from UNK-VQA. As their ambiguous type is when "It has multiple plausible answers", but could be finite possible answers. VLMs can answer by simply listing all.} 
%\lj{Another comment is UNK VQA also contain answerable questions.}

\textbf{Hallucination-based benchmarks: }
MMHal-Bench \cite{DBLP:journals/corr/abs-2309-14525} contains 96 questions curated based on the expert observations in 8 hallucination categories such as object attribute, adversarial object, counting etc., Upon establishing the severity of object hallucinations, \cite{DBLP:conf/emnlp/LiDZWZW23} introduce POPE with $\sim 9K$ instances that samples objects randomly, adversarially, and based on popularity to check for their presence binarily. 
To comprehensively study types of hallucinations, \cite{DBLP:journals/corr/abs-2311-07397} introduce AMBER for existence, attribute, and relation hallucinations and AMBER-based evaluation metrics.

\textbf{Standard benchmarks: } While mitigating hallucination and learning to refuse is important, the goal is also to not hurt model performance on standard datasets. Therefore, we conduct evaluations on standard datasets VQAv2 \cite{DBLP:journals/ijcv/GoyalKASBP19}
\footnote{We randomly sample 5k questions from validation set that is not covered in LLaVA instruction-tuning data.} 
and VizWiz \cite{DBLP:conf/cvpr/Gurari0SGLGLB18} validation splits.

\begin{table*}[!tp]\centering

\tablestyle{9pt}{1.2} 
\def \w{20pt} 
\resizebox{1.0\textwidth}{!}{
\begin{tabular}{lccccccccc}
\shline
&\multicolumn{3}{c}{Epistemic} &\multicolumn{3}{c}{Aleatoric} &\multicolumn{3}{c}{Total} \\
\cmidrule(lr){2-4}
\cmidrule(lr){5-7}
\cmidrule(lr){8-10}
&\multicolumn{2}{c}{$\text{LAVE}_{\text{idk}}$ Metric} &Conf-w. &\multicolumn{2}{c}{$\text{LAVE}_{\text{idk}}$ Metric} &Conf-w. &\multicolumn{2}{c}{$\text{LAVE}_{\text{idk}}$ Metric} &Conf-w. \\
\cmidrule(lr){2-3}
\cmidrule(lr){5-6}
\cmidrule(lr){8-9}
&F1$_{\text{idk}}$ & Acc. &Acc. &F1$_{\text{idk}}$ &Acc. &Acc. &F1$_{\text{idk}}$ &Acc. &Acc. \\
\cmidrule(lr){2-4}
\cmidrule(lr){5-7}
\cmidrule(lr){8-10}
Qwen-VL-Chat &65.45 &64.22 &11.92 &\textbf{67.15} &\textbf{63.35} &15.71 &66.13 &63.87 &13.45 \\
\cmidrule{1-10}
LLaVA-1.5-7B &51.31 &44.72 &-1.01 &54.78 &51.51 &2.65 &52.71 &47.46 &0.47 \\
% LLaVA-1.5-7B-LORA &53.48 &47.30 &1.56 &53.78 &50.55 &0.78 &53.60 &48.61 &1.25 \\
LLaVA-1.5-13B &52.38 &46.14 &2.70 &53.35 &50.46 &1.81 &52.78 &47.88 &2.34 \\
% LLaVA-v1.5-13B-LORA &52.77 &46.59 &6.46 &55.57 &52.46 &5.70 &53.90 &48.95 &6.15 \\
\cmidrule{1-10}
LLaVA-1.6-7B &67.61 &53.10 &26.61 &51.27 &55.51 &11.39 &61.02 &54.07 &20.47  \\
LLaVA-1.6-13B &69.72 &66.88 &28.07 &54.61 &56.72 &14.29 &63.63 &62.78 &22.52 \\
LLaVA-1.6-34B &\textbf{74.37} &\textbf{71.06} &\textbf{40.03} &58.47 &60.01 &\textbf{21.27} &\textbf{67.96} &\textbf{66.60} &\textbf{32.47} \\
\cmidrule{1-10}
\demph{GPT-4V$^\dagger$}&  \demph{85.34} &\demph{78.60} &\demph{-} &\demph{61.41} &\demph{61.25} &\demph{-} &\demph{75.76} &\demph{71.70} & \demph{-}\\
% &\demph{86.90} &\demph{78.60} &\demph{-} &\demph{68.37} &\demph{61.25} &\demph{-} &\demph{79.50} &\demph{71.70} &\demph{-} \\
\shline
\end{tabular}
}
\caption{\small Evaluating existing VLMs on \ourdata. For epistemic/aleatoric category, we average the score across the 3/2 fine-grained categories. Total performance is averaged across all 5 fine-grained categories. $\dagger$ GPT-4V performance are reported on a smaller subset with 100 samples for each finegrained category. 
% \yejin{a bit strange to refer to our 'metric' as 'ours'... (instead of the metric name) because 'ours' usually correlates with a 'method'... even more confusing in Table 3 when 'ours' appear in both columns and rows} 
% \lj{Removed the llava-lora rows, cuz they are repeated in table 3, can remove more rows of llava if no space, and move to Appendix.}  
}\label{tab:benchamrking}
% $\dagger$ GPT-4V performance are reported on a smaller subset, containing only 500 examples, 100 for each finegrained category
% Existing VLMs (even GPT-4V) are doing badly on saying ``I don't know''. 
\end{table*}
\begin{figure}[t]
    \centering
    % \vspace{-3em}
    % \vspace{4pt}
    \includegraphics[width=.98\textwidth, clip]{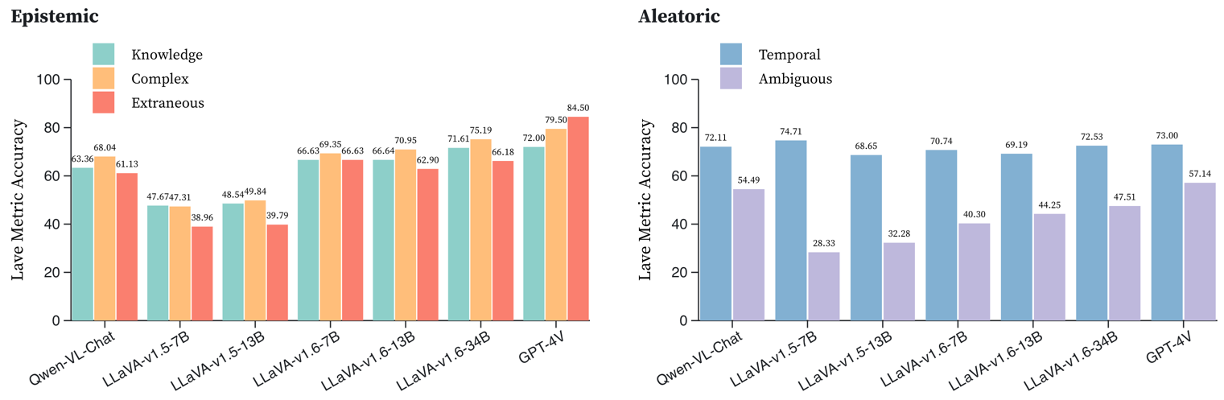}
    \caption{Breakdown of model performance on finegrained categories. We report $\text{LAVE}_{\text{idk}}$ Metric Accuracy as the confidence of GPT-4V prediction is not accessible. 
    % GPT-4V performance are reported on a smaller subset with 100 samples for each finegrained category. 
    % Overall, the average LAVE metric accuracy of GPT-4V is 57.90. 
    % \lj{Did not update yet after resfusal 0/1, will do that after table 3/4.}
    % \lj{Summary of results. Aleatoric: All models are doing badly on ambiguous, temporal seems to be the easiest on average. Compared to Aleatoric, performance difference are slightly smaller among the categories of epistemic.}
    }
    \label{fig:finegrained breakdown}
\end{figure}

\begin{table*}[!tp]\centering

\tablestyle{3pt}{1.2} 
\def \w{20pt} 
\resizebox{1.0\textwidth}{!}{
\begin{tabular}{llcccccccccc}\shline
& &\multicolumn{3}{c}{Epistemic} &\multicolumn{3}{c}{Aleatoric} &\multicolumn{4}{c}{Total} \\
\cmidrule(lr){3-5}
\cmidrule(lr){6-8}
\cmidrule(lr){9-12}
& &\multicolumn{2}{c}{$\text{LAVE}_{\text{idk}}$ Metric} & Conf-w. &\multicolumn{2}{c}{$\text{LAVE}_{\text{idk}}$ Metric} &Conf-w. &\multicolumn{2}{c}{$\text{LAVE}_{\text{idk}}$ Metric} &Conf-w. &ECE $\downarrow$ \\
\cmidrule(lr){3-4}
\cmidrule(lr){6-7}
\cmidrule(lr){9-10}
& &F1$_{\text{idk}}$ &Acc. &Acc. &F1$_{\text{idk}}$ &Acc. &Acc. &F1$_{\text{idk}}$ &Acc. &Acc. &(IDK) \\
\cmidrule(lr){3-5}
\cmidrule(lr){6-8}
\cmidrule(lr){9-12}
\multicolumn{2}{l}{Qwen-VL-Chat$^\star$} &65.45 &64.22 &11.92 &67.15 &63.35 &15.71 &66.13 &63.87 &13.45 &0.79 \\
\cmidrule{1-12}
\multicolumn{2}{l}{Thresholding} &74.44 &69.86 &19.45 &71.84 &62.54 &17.04 &73.40 &66.91 &18.48 &0.62 \\
\cmidrule{1-12}
\multirow{4}{*}{LoRA-SFT} &LRV &57.30 &57.51 &6.69 &50.74 &53.41 &0.42 &54.65 &55.86 &4.16 &0.73 \\
&LLaVA Data &62.18 &62.65 &11.61 &60.81 &62.55 &18.61 &61.63 &62.61 &14.44 &0.68 \\
&Ours &84.62 &76.70 &\textbf{45.00} &86.76 &81.38 &\textbf{55.81} &85.48 &78.59 &\textbf{49.35} &\textbf{0.31} \\
&Ours+LLaVA &\textbf{85.38} &\textbf{78.14} &42.49 &\textbf{87.19} &82.11 &55.27 &\textbf{86.11} &\textbf{79.74} &47.64 &0.37 \\
\hline
\multirow{3}{*}{LoRA-Rtune} &LLaVA Data &69.68 &66.53 &14.92 &73.85 &67.95 &20.78 &71.36 &67.11 &17.28 &0.75 \\
&Ours &\textbf{86.10} &\textbf{78.09} &41.88 &\textbf{85.38} &\textbf{78.52} &51.07 &\textbf{85.81} &\textbf{78.26} &45.58 &0.37 \\
&Ours+LLaVA &85.46 &77.14 &\textbf{44.59} &85.25 &78.20 &\textbf{52.90} &85.37 &77.57 &\textbf{47.94} &\textbf{0.29} \\
\cmidrule{1-12}
\multirow{3}{*}{LoRA-DPO} &MMInstruction &66.10 &65.18 &18.03 &55.98 &56.10 &7.57 &62.02 &61.52 &13.81 &\textbf{0.70} \\
&Ours &\textbf{74.70} &\textbf{69.79} &18.81 &\textbf{73.70} &\textbf{68.59} &\textbf{20.38} &\textbf{74.30} &\textbf{69.30} &\textbf{19.44} &0.78 \\
&Ours+MMinstruction &71.52 &68.46 &\textbf{19.70} &68.51 &64.73 &14.60 &70.31 &66.95 &17.65 &0.75 \\
\shline
\multicolumn{2}{l}{LLaVA-1.5-7B-LoRA$^\star$} &33.72 &37.36 &17.46 &4.59 &50.55 &0.78 &35.11 &48.61 &1.25 &0.62 \\\cmidrule{1-12}
\multirow{2}{*}{Instruct-Tune} &Ours &84.40 &78.25 &\textbf{53.54} &42.07 &81.32 &\textbf{50.33} &85.31 &78.25 &\textbf{42.50} &\textbf{0.41} \\
&Ours+LLaVA & \textbf{85.47} &\textbf{79.60} &46.16 &\textbf{42.57} &\textbf{81.95} &37.62 &\textbf{86.09} &\textbf{79.46} &31.92 &0.64 \\
% \cmidrule{1-12}
% \multicolumn{2}{l}{LLaVA-1.5-13B-LoRA$^\star$} &31.40 &36.08 &19.43 &6.87 &52.46 &5.70 &35.21 &48.95 &6.15 &0.47 \\\cmidrule{1-12}
% \multirow{2}{*}{Instruct-Tune} &Ours &84.73 &78.67 &54.21 &42.02 &81.55 &49.96 &85.57 &78.65 &44.47 &0.38 \\
% &Ours+LLaVA & & & & & & & & & & \\
\shline
\end{tabular}
}
\caption{Comparison on different training strategies with our \ourdata. The best performances are highlighted with bold for each finetuning strategy.
% , numbers in bold indicate it is better than the official released model$^\star$, otherwise, we highlight them with underline. 
Acc: Accuracy. Conf-w.: Confidence-weighted. ECE: Expected Calibration Error. 
%\khyathi{1. SFT is best, DPO is not as effective 2. our data (+llava) is better combo 3. SFT with Our data helps on reducing ECE (IDK), which echoes the title saying idk confidently.}
}\label{tab:finetune}
% $\dagger$ GPT-4V performance are reported on a smaller subset, containing only 500 examples, 100 for each finegrained category
% Existing VLMs (even GPT-4V) are doing badly on saying ``I don't know''. 
\end{table*}

\begin{table*}[!t]\centering

\tablestyle{6pt}{1.2} 
\def \w{20pt} 
\resizebox{1.0\textwidth}{!}{
\begin{tabular}{lcccccccccc}\toprule
&\multicolumn{2}{c}{Refusal } &\multicolumn{4}{c}{Hallucination} &\multicolumn{2}{c}{Standard} \\\cmidrule{4-7}
& \multicolumn{2}{c}{($\text{LAVE}_{\text{idk}}$ Acc. $\uparrow$)} &\multicolumn{2}{c}{MM-Hal } &POPE &\multicolumn{1}{c}{AMBER} & \multicolumn{2}{c}{(VQA score $\uparrow$)}  \\
\cmidrule(lr){2-3}\cmidrule(lr){4-5}\cmidrule(lr){6-6}\cmidrule(lr){7-7}\cmidrule(lr){8-9}
&UNK-VQA &TDIUC &Overall$\uparrow$ &Hall. \% $\downarrow$ &F1 $\uparrow$ &F1 $\uparrow$  &VizWiz &VQAv2 \\
\cmidrule(lr){2-2}\cmidrule(lr){3-3}\cmidrule(lr){4-5}\cmidrule(lr){6-6}\cmidrule(lr){7-7}\cmidrule(lr){8-8}\cmidrule(lr){9-9}
Qwen-VL-Chat &41.32 &95.10 &\textbf{2.89} &0.41 &81.30 &\textbf{87.70 } &66.85 &72.96 \\
LoRA-SFT-LLaVA Data &38.01 &93.18 &2.83 &0.42 &85.61 &86.80  &65.61 &77.26 \\
LoRA-SFT-Ours-only &\textbf{60.65} &\textbf{99.64}&2.70 &\textbf{0.38} &\textbf{86.31 }&81.30 &\textbf{68.40} &69.77 \\
LoRA-SFT-Ours+LLaVA & 59.70 &99.20 &2.75 &0.39 &85.78 &85.90 &67.44 &\textbf{77.32} \\\hline
\multicolumn{9}{l}{\textit{Instruct-tuning}} \\\hline
LLaVA-1.5-7B-LoRA &36.57 &47.36 &2.56 &0.51 &86.06 &84.60  & 51.87 &76.94 \\
Ours-only & 47.71 &95.36 &2.43 &0.47 &73.62 &78.80  &54.10 &49.95 \\
Ours+LLaVA Data &\textbf{49.12} &\textbf{98.70} &\textbf{2.66} &\textbf{0.45} &\textbf{88.05} &\textbf{86.60}  &\textbf{54.40} &\textbf{77.37} \\
% \midrule
% LLaVA-1.5-13B-LoRA &36.91 &84.28 &2.54 &0.51 &85.92 & 85.80	& 6.80 &61.97 &\textbf{77.98} \\
% Ours-only &48.72 &97.42 &2.49 &0.45 &77.04 & 82.00	& 4.10 &58.21 &51.71 \\
% Ours+LLaVA Data & & & \textbf{2.64} &\textbf{0.49} &  & \textbf{86.80} & \textbf{6.60} & &  77.35 \\
\bottomrule
\end{tabular}
}
\caption{Results of different model variants trained with \ourdata on other benchmarks. Hall. \%: Hallucination ratio. 
% Acc: Accuracy. 
$\uparrow$ ($\downarrow$) indicates the larger (smaller) the better.
% \lj{updating vizwiz numbers and UNK-vqa/TDIUC for Qwen. maybe we should remove amber?}  
}\label{tab:others}
\end{table*}

\subsection{Results and Discussion}

We extensively evaluate the performance of GPT-4V, LLaVA and Qwen-VL models on \ourdata. As shown in Table \ref{tab:benchamrking}, we observe that these models including GPT-4V (despite the questions generated with it) perform poorly on our benchmark.  It is also worth noting that all models achieve significantly higher scores on $\text{LAVE}_{\text{idk}}$ accuracy than confidence-weighted accuracy. This discrepancy suggests that the models are either over-confident in incorrect predictions or not confident enough in correct ones (\textit{i.e.}, they are poorly calibrated), which we further examine in our fine-tuning experiments. Figure \ref{fig:finegrained breakdown} presents $\text{LAVE}_{\text{idk}}$ accuracy on each sub-category. The relative trends across sub-categories are consistent among the models. The extraneous category is the most challenging within epistemic uncertainty, while the ambiguous category is the hardest within aleatoric uncertainty. Performance on the temporal category is relatively similar across different model sizes, possibly due to the limited diversity of questions that can be asked about the future.

A comprehensive empirical comparison of the different training strategies is presented in Table \ref{tab:finetune}. For the instruction-tuned Qwen-VL-Chat, we explore different continued finetuning methods with LoRA \cite{DBLP:journals/corr/abs-2309-14717}. For LLaVA, given the availability of their instruct-tuning data, we explore adding our data into instruct-tuning stage. Overall, we observe that SFT learns IDK better on our benchmark compared to other strategies, resulting in higher confidence-weighted accuracy. Within each training strategy—SFT, R-tuning, and DPO—we find that training on our data consistently improves performance, underscoring the quality of our dataset. Finally, SFT with our data also reduces ECE, demonstrating that models trained with \ourdata can express IDK more confidently.

Lastly, we examine the performance of our finetuned/instruction-tuned models on other benchmarks in Table \ref{tab:others}. The results show that our dataset effectively improves model performance on refusal-based benchmarks, including UNK-VQA and TDIUC. It also demonstrates promising trends in reducing hallucination ratios in MM-Hal and improving F1 scores on POPE. Despite \ourdata focusing solely on VQA-type data, when finetuned/instruction-tuned with our data only, we did not observe a significant drop in the overall score of MM-Hal, which also evaluates tasks like captioning. When augmentating the LLaVA instruction tuning data with ours, it even improves the overall score of MM-Hal for LLaVA model.
On AMBER, for Qwen-VL-Chat, SFT with any data combination in our experiments led to inferior results, especially when using only our data. We speculate that the degradation on AMBER is due to the lack of IDK questions on attributes and relations about non-existent objects in our dataset, which we plan to extend upon in future work. In comparison, POPE mainly focuses on existential questions about objects, which is more similar to our extraneous split. Moreover, on standard VQA benchmarks, we observe that models trained with our data combined with LLaVA data perform on par with the VQAv2 benchmark and show improvements on VizWiz, which contains unanswerable questions.

% both consisting of 
% % Both POPE and AMBER feature binary questions. POPE focuses on o

% existential questions. 
% <types of POPE questions closer to our data>
% <difference with AMBER>.
% \khyathi{talk about table 5}

\section{Related Work}

\textbf{Abstention. } 
Early studies in abstention primarily focused on the notion of confidence estimation in predictions, allowing 
to abstain when uncertain \cite{chow1957optimum, de2000reject, el2010foundations}. Recent works used selective prediction approaches to particularly improve reliability under domain shift \cite{DBLP:conf/acl/KamathJL20} and with adversarial inputs \cite{varshney2022investigating}. Another promising direction involves extracting additional evidences 
by iteratively accumulating context \cite{DBLP:journals/corr/abs-2402-15610, zeng2022socratic, shen2023hugginggpt, you2023idealgpt, yang2023mm}, rephrasing underspecified questions \cite{prasad2023rephrase}, 
probing through code \cite{gupta2023visual, surismenon2023vipergpt, subramanian2023modular}. Unlike our work, these approaches aim to reduce the risk of incorrect predictions despite having definitive answers, without addressing epistemic or aleatoric uncertainties.

\textbf{Hallucinations.} Models tend to over-confidently hallucinate in uncertain scenarios \cite{DBLP:journals/corr/abs-2403-04307, DBLP:journals/corr/abs-2311-09114, DBLP:journals/corr/abs-2310-16045}. 
There are two primary techniques for hallucination detection \cite{DBLP:journals/corr/abs-2401-08358} -- at token-level \cite{Liu2021ATR, Zhou2020DetectingHC, Dziri2021NeuralPH, Cao2021HallucinatedBF} and sentence-level \cite{Manakul2023SelfCheckGPTZB, Zha2023AlignScoreEF, Shen2023WhyIT, Li2023HaluEvalAL}. We aim to reduce the confidence of hallucinatory responses at the answer-level. 
Similar to extracting additional evidence, hallucination mitigation strategies use retrieval-based approaches \cite{Ji2022RHOR, Dziri2021NeuralPH, Shuster2021RetrievalAR}, which condition outputs on factual data by using external knowledge sources, particularly helps increase reliability on the \textit{extraneous} category.

\textbf{Evaluation. } Standard accuracy or generation metrcics such as BLEU are insufficient to evaluate the confidence of open-ended answer generation. To assess the semantic possibilities the LAVE metric \cite{DBLP:conf/aaai/ManasKA24}  was introduced to fully or partially score the predicted answer based on their overlap with the ground truth. Expected Calibration Error (ECE) measures the accuracy of probability estimates in representing true correctness likelihood. More recent approaches also rely on object detection \cite{Johnson2015DenseCapFC, Rohrbach2018ObjectHI, Li2023EvaluatingOH, Lovenia2023NegativeOP, Gunjal2023DetectingAP} or entailment (Faithscore) \cite{DBLP:journals/corr/abs-2311-01477} to measure hallucinations. However, none of these metrics directly indicate the confidence of the model predictions. In this work, we build upon LAVE accuracy by introducing confidence-weighted accuracy, which better correlates with ECE.
% So we build upon the LAVE accuracy and introduce confidence-weighted accuracy that is better correlated with ECE as well.

\textbf{Datasets. }
Most standard multimodal benchmarks focus on clear, definitive answers or partial hallucinations  \cite{DBLP:conf/aaai/GunjalYB24, DBLP:journals/corr/abs-2404-18930} for discriminative  \cite{Li2023EvaluatingOH, DBLP:journals/corr/abs-2311-16479, DBLP:conf/mmm/WangHLLL24} or generative tasks \cite{liu2023mitigating, DBLP:journals/corr/abs-2311-01477}. In contrast, \ourdata targets scenarios where being underconfident or responding with IDK is the correct response. 
Similar concurrent efforts for text-only benchmarks  \cite{Li2023HaluEvalAL, Zhu2024HaluEvalWildEH, Mishra2024FinegrainedHD} are widely explored. Generating counterfactual instruction text data \cite{DBLP:journals/corr/abs-2311-13614} is the closest equivalent of LRV data \cite{liu2023mitigating} which includes positive (or negative) instructions about objects or attributes present (or absent) in the image. We also use the MMInstruction \cite{DBLP:journals/corr/abs-2312-10665} with preference annotations for helpfulness, faithfulness and ethical considerations. Finally, we generate model-dependent refusal datasets automatically which is explored by \cite{Zhang2023RTuningIL} to adapt to multimodal R-tuning. Our experiments show that these datasets are insufficient for benchmarking or improving multimodal epistemic and aleatoric awareness.

\paragraph{}

% \vspace{-1em}
\section{Conclusions and Future Work}
% \vspace{-1em}

% Acknowledging uncertainty in responses and appropriately responding with ``I don't know" (IDK) is crucial for the reliability and trustworthiness of VLMs.
% %Our work highlights the inadequacy of current VLMs in this regard and 
% We introduce a new taxonomy specifically designed to handle epistemic and aleatoric awareness in multimodal uncertainty. To this end, we present a new benchmarking dataset \ourdata and demonstrate that VLMs are not self-aware of uncertainties. We also empirically demonstrate the effectiveness of this benchmark by showing performance gains when different models with different training strategies are fine-tuned on our data --  particularly showing improvements on hour held-out test set, existing refusal-based benchmarks and some improvements on hallucination-based benchmarks while maintaining the performance on standard benchmarks. Additionally, we propose a new confidence-weighted accuracy metric that combines evaluating the predictive performance with the confidence of the prediction, which is well correlated with both accuracy and ECE. This work paves the way for numerous future research opportunities. We plan to extend the annotations to provide rationales about which category is the reason for saying IDK. Our confidence-weighted metric holds potential for exploration in other unimodal and multimodal datasets, and it can also be investigated as a reward mechanism.

Acknowledging uncertainty in responses and appropriately responding with ``I don't know" (IDK) is crucial for the reliability and trustworthiness of VLMs. In this work, we introduce a new taxonomy specifically designed to handle epistemic and aleatoric uncertainty in multimodal systems. Based on this taxonomy, we present a new benchmarking dataset, \ourdata, and demonstrate that current VLMs lack self-awareness of these uncertainties.
Empirical results show that fine-tuning  with our data leads to performance gains, particularly on our held-out test set, existing refusal-based benchmarks, and some hallucination-based benchmarks, all while maintaining performance on standard benchmarks. Additionally, we propose a new confidence-weighted accuracy metric that combines predictive performance with the confidence of the prediction, showing strong correlations with both accuracy and ECE.
Our work paves the way for future research directions in modeling uncertainties. For instance, future efforts could extend the annotations to include rationales for IDK responses, specifying which category of uncertainty is responsible. Additionally, our confidence-weighted metric holds potential for application in other unimodal and multimodal datasets and could be explored as a reward mechanism in model training.

%\khyathi{future work: 1. extending annotations to rationale; 2. using accuracy as metric}

% these foundations, exploring new strategies and refining evaluation metrics to further enhance the reliability and trustworthiness of LLMs.

% \input{sections/repro_statement.tex}
% \input{sections/acknowledgements.tex}

\bibliographystyle{unsrt}
\bibliography{references}
\newpage

\newpage
\appendix

% \paragraph{Prompts}

\section{Limitations}
\label{sec:limitations}

While our \ourdata covers various categories of multimodal uncertainty, and showed improvements over the base model when finetuned with it, there are potential limitations to be acknowledged. 
Though our synthetic data is rigorously quality-checked, 
it is possible that the synthetic generation pipeline may not capture all the nuances of real-world uncertain scenarios. Additionally, the most effective way to improve model performance on our benchmark currently is SFT with LoRA, which is more resource-intensive compared to techniques such as selective prediction that makes decisions based on the prediction probabilities during inference. Moreover, providing a reasonable or best guess based on existing knowledge can be more suitable than either answering or abstaining, which we leave as future work.

\section{Broader Impact}
Current models are incentivized to predict definitive answers even in uncertain scenarios. This can lead to outputs with unwarranted confidence, which is particularly problematic in high-stakes applications such as medical diagnosis or financial forecasting. This tendency can result in misleading information and erroneous decisions. In critical applications, incorporating uncertainty awareness can significantly enhance safety %used safety deliberately to show broader impact
and trust by highlighting areas where human expertise is essential. Our proposed taxonomy and data creation pipeline can be adapted to various scenarios, provided domain-specific inpainting techniques are available. Additionally, when models are trained with \ourdata, it can facilitate more efficient resource allocation, as models can identify when additional data or analysis is required, ultimately leading to more robust and trustworthy models. Specifically, identifying the category of epistemic and aleatoric awareness from \ourdata can help identify better means to tackle the uncertainty. Finally, our confidence-weighted metric allows for comprehensive performance evaluation across a wide range of domains, encompassing both unimodal and multimodal scenarios.

\section{Samples visualizing \ourdata benchmark}
\label{sec:data}
%We visualize samples of \ourdata for each finegrained category of epistemic awareness in Figure~\ref{fig:epistemic_sample} and aleatoric awareness in Figure~\ref{fig:aleatoric_sample}.

We visualize some samples from each fine-grained category of the epistemic and aleatoric awareness. For the category of \textit{extraneous}, our data is made of samples where the answer differs for the same question when the image is perturbed. For the rest of the categories, the dataset contains samples where the same image is paired with answerable and unanswerable questions.

\begin{figure}[H]
\centering
\begin{minipage}{.5\textwidth}
  \centering
  \captionsetup{width=\linewidth}
  \includegraphics[width=\textwidth, trim={6cm 11cm 32cm 4cm},clip]{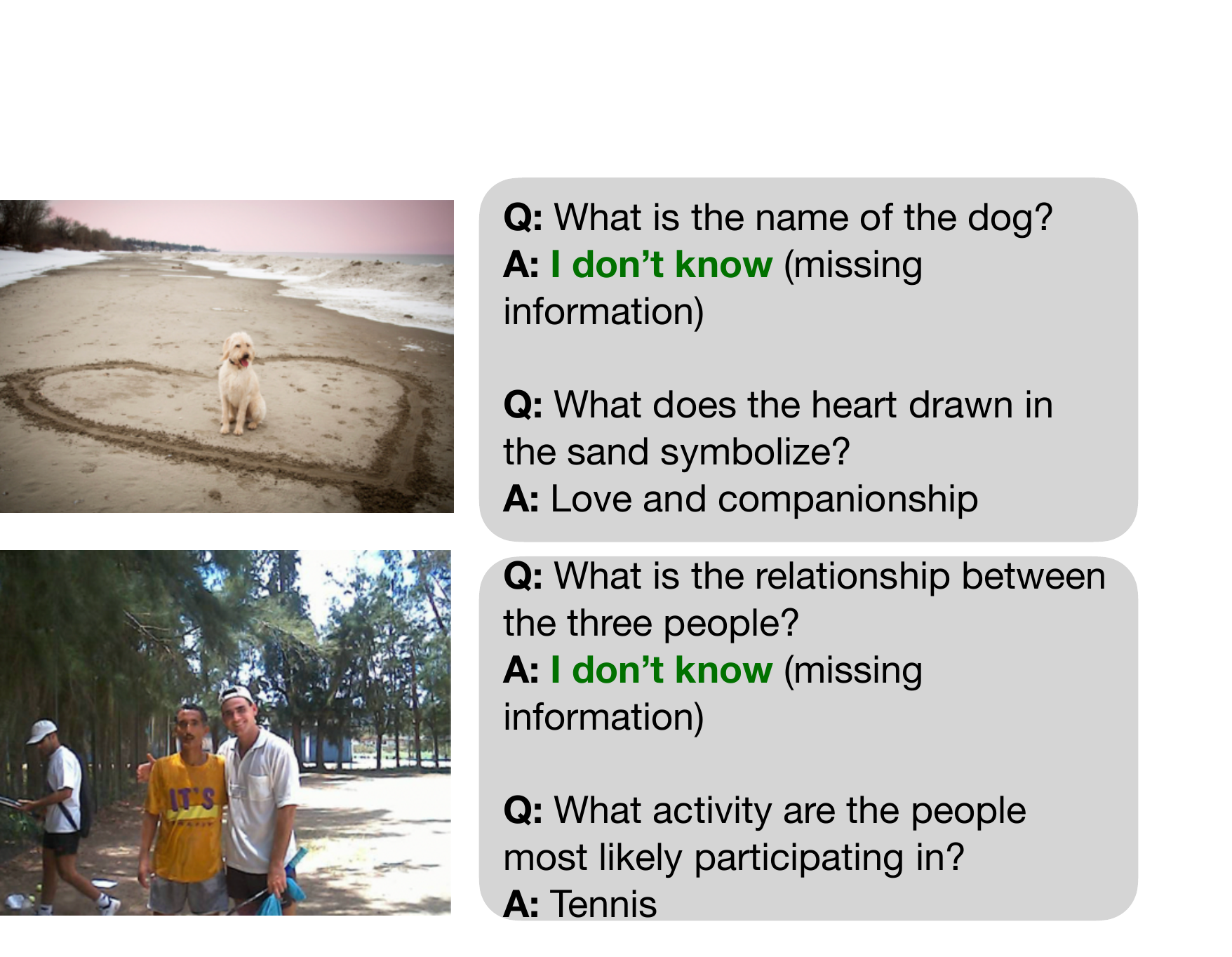}
  \captionof{figure}{Samples from Knowledge Awareness \\ (Epistemic) category}
  \label{fig:knowledge}
\end{minipage}%
\begin{minipage}{.5\textwidth}
  \centering
  \captionsetup{width=\linewidth}
  \includegraphics[width=\textwidth, trim={6cm 10cm 32cm 5cm},clip]{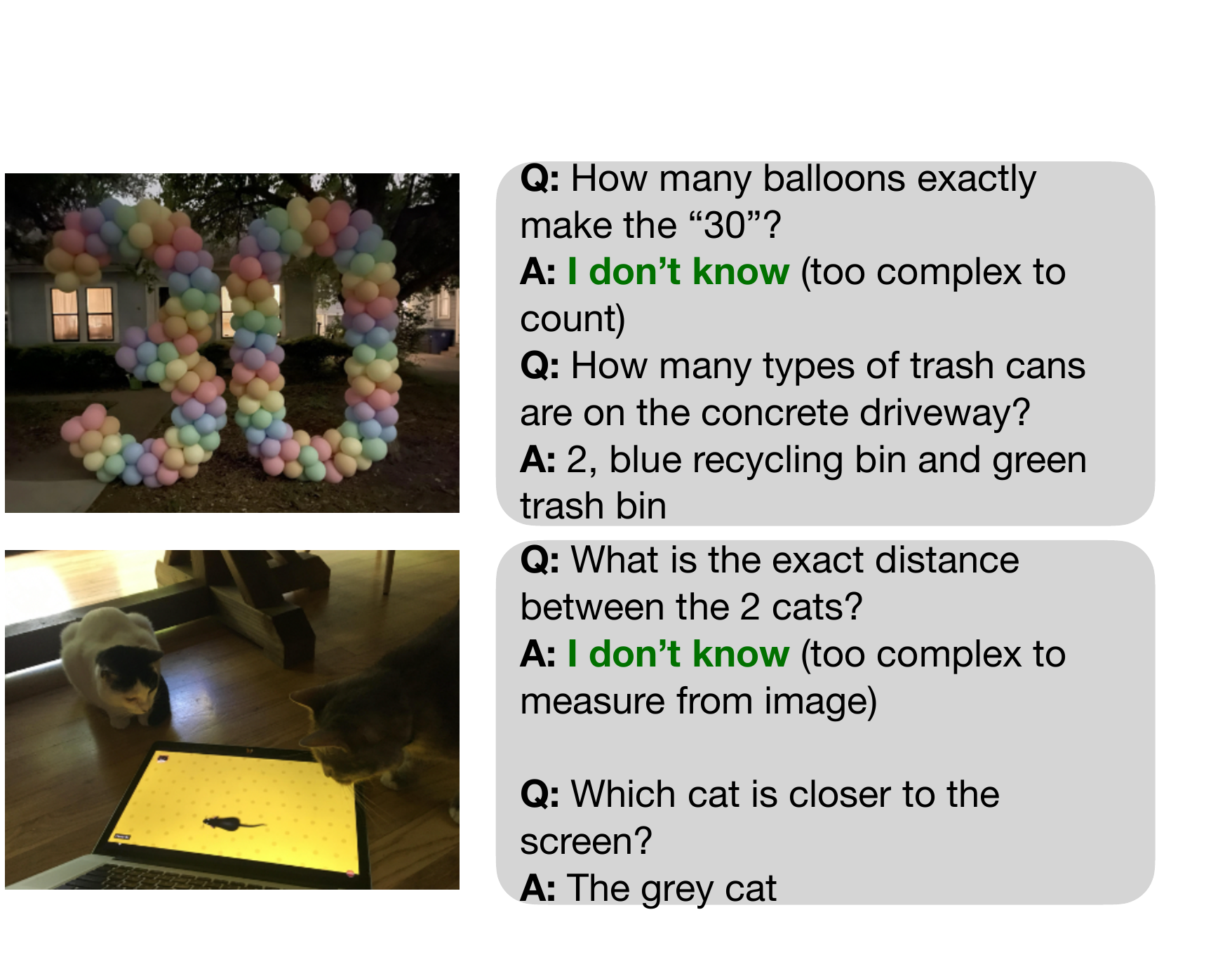}
  \captionof{figure}{Samples from Complexity Awareness \\ (Epistemic) category}
  \label{fig:complex}
\end{minipage}
\end{figure}

Figure \ref{fig:knowledge} shows the category of knowledge awareness; as we can see the unanswerable questions ask about information that is hard to identify from the context of the image and requires additional knowledge. 
Similarly, Figure \ref{fig:complex} shows examples from the complexity awareness in the epistemic category. The unanswerable questions are too tedious to arrive at an answer while the answerable questions still require some efforts, such as counting but is not laborious to answer.

\begin{figure}[H]
  \centering
  \captionsetup{width=\linewidth}
  \includegraphics[width=0.99\textwidth, trim={10cm 19cm 8cm 4cm},clip]{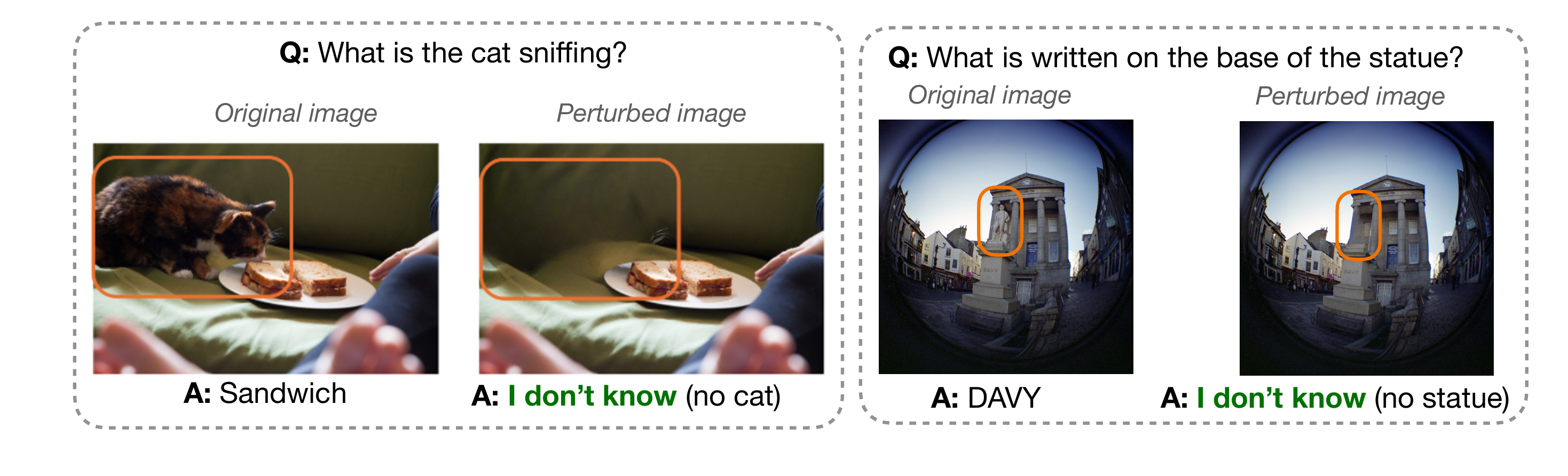}
  \captionof{figure}{Samples from Extraneous Awareness (Epistemic) category}
  \label{fig:extraneous}
\end{figure}

For the extraneous sub-category of the epistemic awareness, we perturb the image to mask and remove the target object about which the question seeks information. Samples from this category are shown in Fig \ref{fig:extraneous}. As we can see the target objects in the questions are `cat' and `statue'. These objects are removed from the image and inpainted to get a natural-looking image to obtain a perturbed image. The resulting image paired with the same question now becomes unanswerable. The answerable question is paired with the original unperturbed question to have a definitive answer (which is the standard setup for most VQA based benchmarks).

\begin{figure}[H]
\centering
\begin{minipage}{.5\textwidth}
  \centering
  \captionsetup{width=\linewidth}
  \includegraphics[width=\textwidth, trim={6cm 11cm 32cm 4cm},clip]{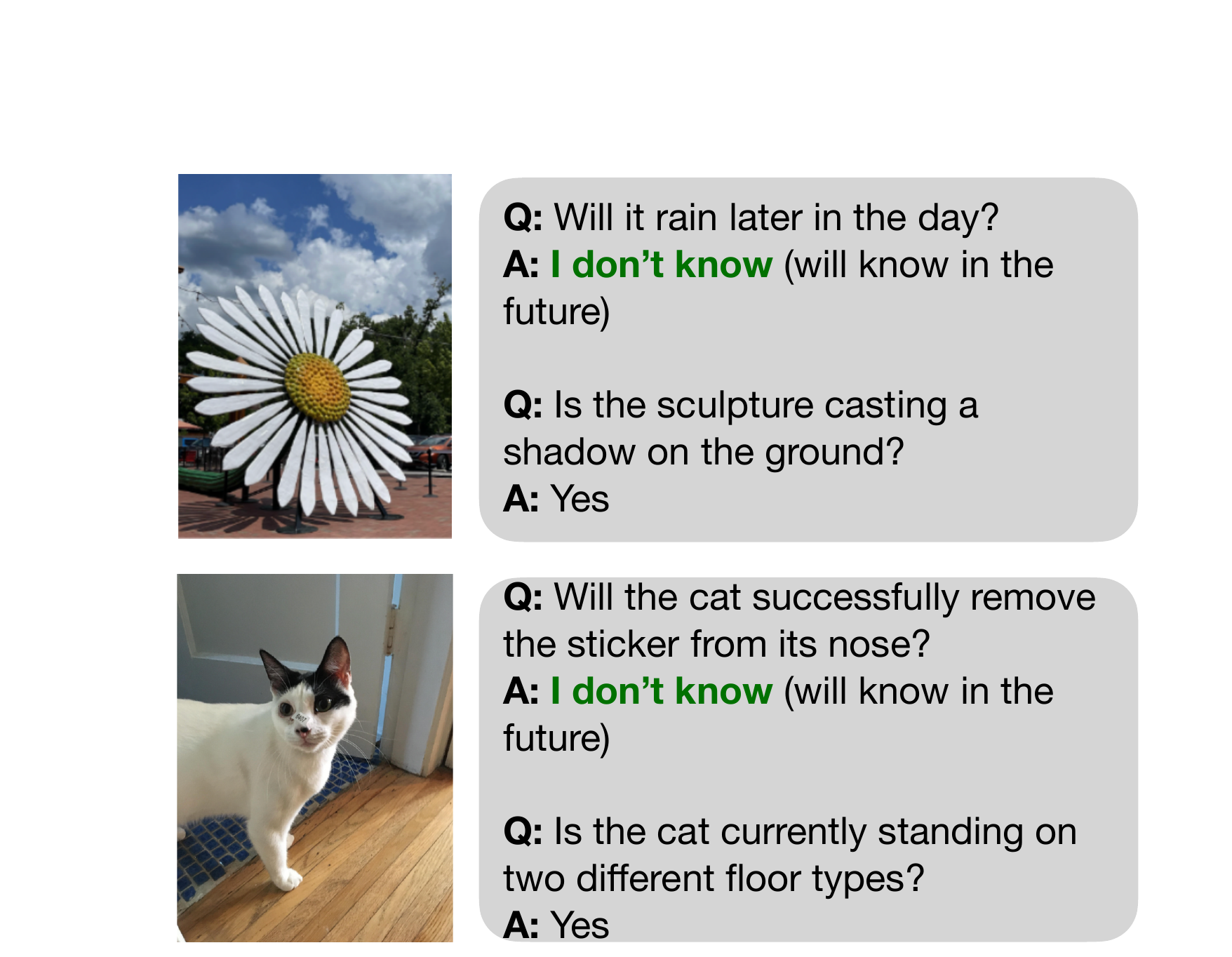}
  \captionof{figure}{Samples from Temporal Awareness \\ (Aleatoric) category}
  \label{fig:temporal}
\end{minipage}%
\begin{minipage}{.5\textwidth}
  \centering
  \captionsetup{width=\linewidth}
  \includegraphics[width=\textwidth, trim={6cm 10cm 32cm 5cm},clip]{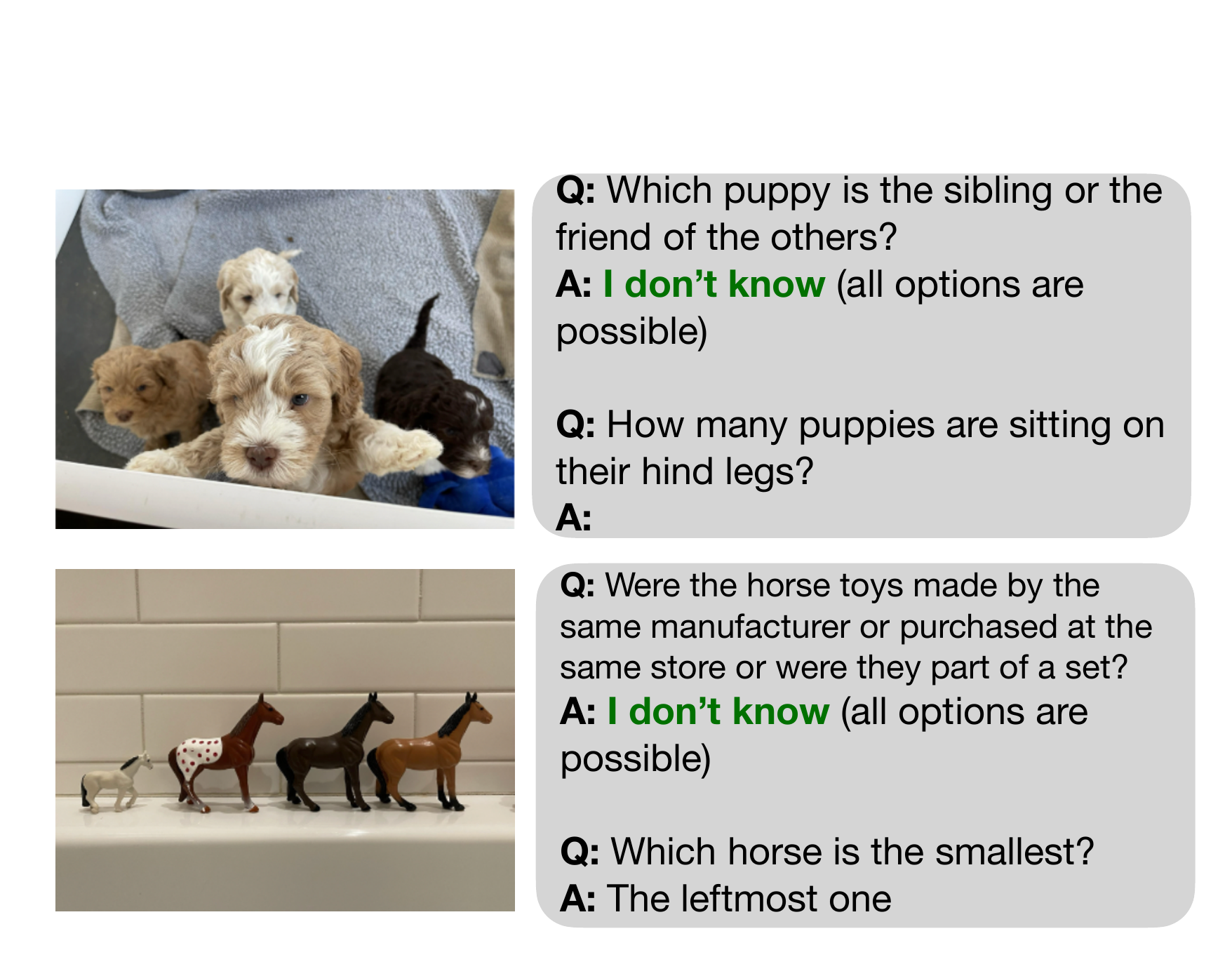}
  \captionof{figure}{Samples from Ambiguous Awareness \\(Aleatoric) category}
  \label{fig:ambiguous}
\end{minipage}
\end{figure}

Figures \ref{fig:temporal} and \ref{fig:ambiguous}
 show samples from \ourdata in the aleatoric category, particularly of the temporal and ambiguity awareness sub-categories respectively. The temporal sub-category, as we can see, contains questions about the current happenings or state of the image for the answerable part and the unanswerable questions ask about the future that is not directly predictable from the current state. 
The ambiguous awareness category contains questions with a definitive answer for the answerable type and questions with many plausible answers but cannot choose a single definite answer for the unanswerable type.

\begin{figure}[H]
  \centering
  \captionsetup{width=\linewidth}
  \includegraphics[width=0.99\textwidth, trim={0cm 0cm 0cm 0cm},clip]{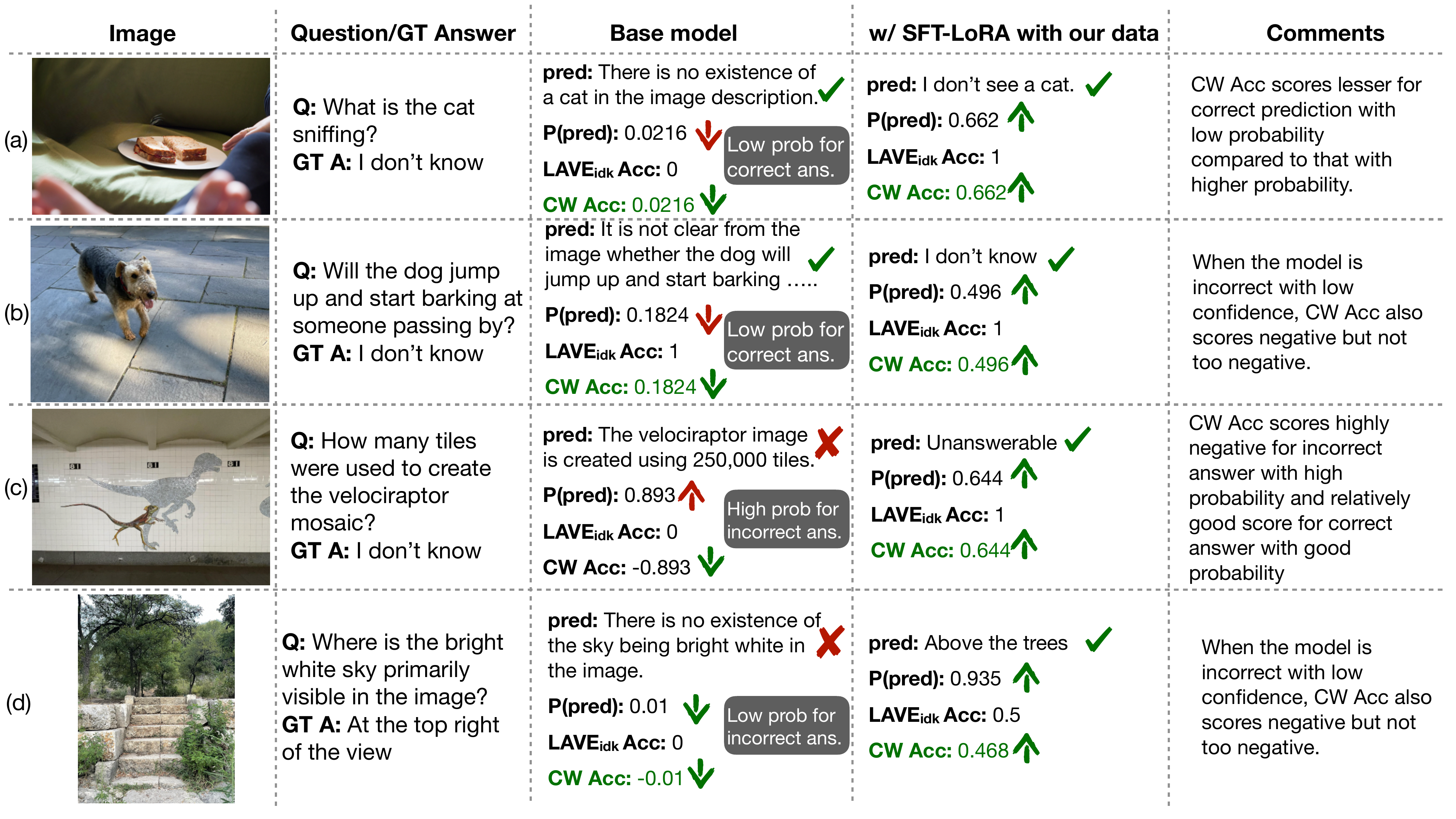}
  \captionof{figure}{Visualization of model predictions and the corresponding $\text{LAVE}_{\text{idk}}$ accuracy, $P_{\text{pred}}$ and confidence-weighted accuracy. The base model here is Qwen-VL-Chat \cite{bai2023qwen}. Our confidence-weighted accuracy is represented as CW Acc in this figure.}
  \label{fig:cw_vis}
\end{figure}

\section{Samples visualizing predictions and \textit{confidence-weighted metric}}
\label{sec:model_pred}

Our proposed confidence-weighted accuracy takes into account the prediction probability and the correctness of the predicted answer to give a holistic score. 
Figure~\ref{fig:cw_vis} presents the visualization of model predictions and the corresponding $\text{LAVE}_{\text{idk}}$ accuracy, $P_{\text{pred}}$ and confidence-weighted accuracy. We show that the proposed confidence-weighted accuracy gives less score for a correct answer with lower confidence, and penalizes more for an incorrect answer with higher confidence. %(Figure~\ref{fig:supp_model_pred2} and~\ref{fig:supp_model_pred3}). 
In addition, our visualization shows that Qwen-VL-Chat~\cite{bai2023qwen} is able to say equivalents of ``I don't know" more confidently from (a) and (b), after continued finetuning on our data with SFT-LoRA. 

Examples (a) and (b) show cases where the base model is less confident for a correct answer. Our metric gives a partial score for the correctness owing to the prediction probability. After finetuning, as the prediction probability of the correct answer increases, our confidence-weighted accuracy increases accordingly. In case (c), the base model predicts an incorrect answer with high confidence. Our metric penalizes this more heavily with a high negative score. After finetuning, the prediction is rectified and the scores are adjusted accordingly. In the case of (d), the base model predicts the incorrect answer but with low confidence. Our metric still gives a negative score but penalizes less compared to (c). The cases of (c) and (d) differentiate answering incorrectly with high and low probabilities respectively.

Moreover, after finetuning with our \ourdata, we see the corrected predictions with relatively higher probabilities for correctness, which are reflected in our confidence-weighted metric score. These probabilities of the model predictions are not reflected in the $\text{LAVE}_{\text{idk}}$ accuracy.

\section{Additional Results}\label{sec:supp_add_results}
% \subsection{More results on LLaVA}

\textbf{LLaVA with LoRA-SFT.} We include results with LoRA-SFT on LLaVA-v1.5-7b in Table~\ref{tab:app_llava}, which show consistent performance improvement when trained with our data. 
\begin{table*}[!htp]\centering
\tablestyle{8pt}{1.0}

\resizebox{1.0\textwidth}{!}{

\begin{tabular}{llrrrrrrrrrr}\toprule
& &\multicolumn{3}{c}{Epistemic} &\multicolumn{3}{c}{Aleatoric} &\multicolumn{3}{c}{Total} \\
\cmidrule(lr){3-5}
\cmidrule(lr){6-8}
\cmidrule(lr){9-11}
& &\multicolumn{2}{c}{$\text{LAVE}_{\text{idk}}$ Metric} & Conf-w. &\multicolumn{2}{c}{$\text{LAVE}_{\text{idk}}$ Metric} &Conf-w. &\multicolumn{2}{c}{$\text{LAVE}_{\text{idk}}$ Metric} &Conf-w.   \\
\cmidrule(lr){3-4}
\cmidrule(lr){6-7}
\cmidrule(lr){9-10}
& &F1$_{\text{idk}}$ &Acc. &Acc. &F1$_{\text{idk}}$ &Acc. &Acc. &F1$_{\text{idk}}$ &Acc. &Acc. \\
\cmidrule(lr){3-5}
\cmidrule(lr){6-8}
\cmidrule(lr){9-11}
\multicolumn{2}{l}{LLaVA-v1.5-7b} &30.08 &44.72 &-1.01 &42.38 &53.39 &8.54 &33.77 &47.46 &0.47 \\\cmidrule{1-11}
\multirow{2}{*}{LoRA-SFT} &Ours &\textbf{85.57} &\textbf{77.83} &\textbf{30.80} &\textbf{84.85} &\textbf{78.80} &\textbf{30.55} &\textbf{86.20} &\textbf{79.53} &\textbf{31.68} \\\cmidrule{2-11}
&Ours+LLaVA data &85.13 &77.14 &21.96 &84.28 &78.12 &20.70 &85.73 &78.85 &26.53 \\
\bottomrule
\end{tabular}
}
\caption{\small Results of LoRA-SFT with LLaVA-v1.5-7b on \ourdata. The best performances are highlighted with bold.}\label{tab:app_llava}
\end{table*}

\textbf{Comparing 7B to 13B models.}
% with LLaVA-1.5-13B, with or without LoRA}
% \label{sec:supp_add_results}
We conduct experiments to study the performance of a larger model across different uncertainty awareness categories. These results are presented in Table~\ref{tab:supp-instruct-tune}.

\begin{table*}[!tp]\centering

\tablestyle{4pt}{1.2} 
\def \w{20pt} 
\small
\resizebox{1.0\textwidth}{!}{
\begin{tabular}{llcccccccccc}\shline
& &\multicolumn{3}{c}{Epistemic} &\multicolumn{3}{c}{Aleatoric} &\multicolumn{4}{c}{Total} \\
\cmidrule(lr){3-5}
\cmidrule(lr){6-8}
\cmidrule(lr){9-12}
& &\multicolumn{2}{c}{$\text{LAVE}_{\text{idk}}$ Metric} & Conf-w. &\multicolumn{2}{c}{$\text{LAVE}_{\text{idk}}$ Metric} &Conf-w. &\multicolumn{2}{c}{$\text{LAVE}_{\text{idk}}$ Metric} &Conf-w. &ECE $\downarrow$ \\
\cmidrule(lr){3-4}
\cmidrule(lr){6-7}
\cmidrule(lr){9-10}
& &F1$_{\text{idk}}$ &Acc. &Acc. &F1$_{\text{idk}}$ &Acc. &Acc. &F1$_{\text{idk}}$ &Acc. &Acc. &(IDK) \\
\cmidrule(lr){3-5}
\cmidrule(lr){6-8}
\cmidrule(lr){9-12}
\multicolumn{2}{l}{LLaVA-1.5-7B-LoRA$^\star$} &33.72 &37.36 &17.46 &4.59 &50.55 &0.78 &35.11 &48.61 &1.25 &0.62 \\\cmidrule{1-12}
\multirow{2}{*}{7B-LoRA-Instruct-Tune} &Ours-only &84.40 &78.25 &\textbf{53.54} &42.07 &81.32 &\textbf{50.33} &85.31 &78.25 &\textbf{42.50} &\textbf{0.41} \\
&Ours+LLaVA Data & \textbf{85.47} &\textbf{79.60} &46.16 &\textbf{42.57} &\textbf{81.95} &37.62 &\textbf{86.09} &\textbf{79.46} &31.92 &0.64 \\
\cmidrule{1-12}
\multicolumn{2}{l}{LLaVA-1.5-13B-LoRA$^\star$} &31.40 &36.08 &19.43 &6.87 &52.46 &5.70 &35.21 &48.95 &6.15 &0.47 \\\cmidrule{1-12}
\multirow{2}{*}{13B-LoRA-Instruct-Tune} &Ours-only &84.73 &78.67 &\textbf{54.21} &42.02 &81.55 &\textbf{49.96} &85.57 &78.65 &\textbf{44.47} &\textbf{0.38} \\
&Ours+LLaVA Data & \textbf{85.99} &\textbf{80.32} &48.50 &\textbf{42.61} &\textbf{82.53} &48.80 &\textbf{86.55} &\textbf{80.20} &42.00 &0.47 \\
\shline
\end{tabular}
}
\caption{Scaling results on instruct tuning LLaVA with the augmentation of \ourdata. $^\star$ indicate we directly load the released weight from LLaVA official implementation. The best performances are highlighted with bold.}\label{tab:supp-instruct-tune}
\vspace{1em}
\end{table*}

We observe consistent performance improvements over LLaVA-1.5-7B-LoRA and LLaVA-1.5-13B-LoRA~\cite{DBLP:conf/nips/LiuLWL23a} with the augmentation of \ourdata during the instruction-tuning phase.  When instruction-tuned with only our data (\textit{i.e.}, Ours-only), compared to the results on the 7B-LoRA model, a larger model 13B-LoRA only marginally improves on confidence-weighted accuracy and ECE (IDK). However, when mixing our data with LLaVA instruction tuning data (\textit{i.e.}, Ours+LLaVA Data), the resulting 13B model clearly outperforms 7B on both metrics. 
% This shows that scaling the model size helps improve the performance of the larger models but is better with increased instruction tuning data in the training mix. 
% While for the smaller 7B model, including LLaVA instruction tuning data in the training mix only improves the performance on LAVE, but does not help with ECE or our confidence-weighted metric.

In addition, we observe that the model performance on $\text{LAVE}_{\text{idk}}$ metrics stay on par for 7B and 13B models with the same training data,
% \khyathi{what strategies},
while they can still be differentiated by our proposed metric, which further highlights the importance of confidence-weighted accuracy.
% \textbf{Scale experiments:}

\begin{table*}[!t]\centering
\tablestyle{8pt}{1.0}
% \footnotesize

\small
\resizebox{.45\textwidth}{!}{
\begin{tabular}{ccccc}\toprule
Mistral-7B &Yi-34B-4bits &GPT4 &Human \\
98\% &100\% &100\% &100\% \\
\bottomrule
\end{tabular}
}
\caption{LAVE evaluator accuracy on GT IDK. Performance are reported on 100 random samples from extraneous split. }\label{tab:lave_refusal}
\end{table*}
\textbf{LAVE IDK Judgement Accuracy} As our metric relies on the accuracy of LAVE refusal judgment, we experiment with different LLMs on 100 samples from extraneous split. Table~\ref{tab:lave_refusal} presents the results with Mistral-7B~\cite{jiang2023mistral}, Yi-34B-4bits~\cite{ai2024yi} and GPT4, in comparison with Human. Given the high performance, and also considering the latency and cost of model inference, we decide to use the smaller model Mistral-7B in our evaluation.

\section{Implementation Details}
\label{sec:experimets_appendix}
% \textbf{Implementation Details.}
For \textit{Thresholding} baselines, we perform grid search among $(0.1, 0.2, ... 0.9)$ and $(0.91, 0.92, ... 0.99)$ to decide the optimal threshold for each split. The latter range is included, as we observe that the models are often over-confident in their own predictions.

For \textit{SFT/Instruction-tuning with LoRA}, we follow the instructions provided by Qwen-VL and LLaVA official implementations, with exactly the same setting of learning rate and LoRA configurations.  For \textit{Rtune}, we construct the dataset by first running inference on the training split of LLaVA data and our dataset, and then gather the instances where the model predicts a wrong answer (\textit{i.e.,} receives a LAVE accuracy of 0). With the constructed dataset, we tune Qwen-VL with the same training configuration as SFT. For \textit{DPO}, we follow the implementations of Silkie~\cite{DBLP:journals/corr/abs-2312-10665}.
% for Qwen-VL-Chat
% , and HA-DPO~\cite{zhao2023hallucinations} for LLaVA.

All experiments are conducted with V100s on Microsoft Azure~\cite{msft-azure}, adopting mixed-precision training with DeepSpeed~\cite{rasley2020deepspeed} stage 3. To match the batch size suggested in official implementations, we train the models on 64 V100s for 1 epoch with a batch size of 2 per GPU.

For evaluation on Vizwiz, we first use LAVE refusal prompt to judge whether the prediction is IDK. If so, we convert the answer to ``unanswerable'' and use the standard VQA-based VizWiz evaluation.

\section{Additional Details on Data Creation}

% \subsection{Samples in \ourdata}

\subsection{More Details on sourcing from image}
The masks of salient objects are generated by Grounded-SAM~\cite{DBLP:journals/corr/abs-2401-14159} with box\textunderscore threshold of 0.3 and text\textunderscore threshold of 0.25. The mask is dilated with kernel size 20 and then input to LaMa inpainting model~\cite{suvorov2021resolution} to remove the object.

For VQA images, we use GPT-4 to first identify the salient objects given the question-answer pairs, which will use as text queries to Grounded-SAM. 

For GQA images, we identify objects in the scene graphs that is associated with a question as the salient object. Then we traverse the scene graphs to find all other objects with the same label. Since GQA also offers groundtruth bounding box (bbox) annotations, we use the mask generated by Grounded-SAM from GT bbox, following by inpainting to remove all such objects. In this way, the same question becomes unanswerable for the perturbed image, and we replace the answer with IDK answers by randomly sample from (1) ``I don't know."; (2) ``I don't see any [Object]."; (3) ``There is no [Object] in the image."; and (4) ``I can't see any [Object].".

\subsection{Prompts for Data Creation}
Here are the prompts for generating data in epistemic and aletoric subcategories with GPT-4 or GPT-4V.
%%%%%%%%%%

\begin{tcolorbox}[colback=cyan!5!white, colframe=cyan!75!black, title=Knowledge (Epistemic Awareness), breakable]
\small{You are given a descriptive caption of an image. Generate a knowledge based answerable and an unanswerable question from the cation.
An unanswerable question requires external knowledge or commonsense that is not explicitly absent in the image to answer the question.
An answerable question requires commonsense knowledge not present in the image pixels but can be answered from the context.

Make the unanswerable and answerable questions as similar to each other as possible yet one is answerable and the other is unanswerable. 
Here are some examples:
\\~\\
Caption: In the center of the image, a vibrant blue lunch tray holds four containers, each brimming with a variety of food items. The containers, two in pink and two in yellow, are arranged in a 2x2 grid. In the top left pink container, a slice of bread rests, lightly spread with butter and sprinkled with a handful of almonds. The bread is cut into a rectangle, and the almonds are scattered across its buttery surface. Adjacent to it in the top right corner, another pink container houses a mix of fruit. Sliced apples with their fresh white interiors exposed share the space with juicy chunks of pineapple. The colors of the apple slices and pineapple chunks contrast beautifully against the pink container. Below these, in the bottom left corner of the tray, a yellow container holds a single meatball alongside some broccoli. The meatball, round and browned, sits next to the vibrant green broccoli florets. Finally, in the bottom right yellow container, there's a sweet treat - a chocolate chip cookie. The golden-brown cookie is dotted with chocolate chips, their dark color standing out against the cookie's lighter surface. The arrangement of these containers on the blue tray creates a visually appealing and balanced meal, with each component neatly separated yet part of a cohesive whole.

Unanswerable Q: How many calories in this meal?

Answer: Unanswerable

Answerable Q: Which cuisine is the meal?

A: English meal 
\\~\\
Caption: This image captures a fascinating scene in a dense jungle. Two majestic, gray elephants are the main subjects of the photo. They are carrying people on their backs, who are seated in wooden seats and wearing helmets for safety. The elephants are walking in a line, one following the other, on a path that cuts through the lush greenery of the jungle. The photo is taken from a higher vantage point, providing a bird's eye view of the elephants and their verdant surroundings. The dense foliage and towering trees of the jungle envelop the path, creating a sense of adventure and exploration.

Unanswerable Question: What are the relationships between the people on the elephants?

Answer: Unanswerable

Answerable Question: Who are the people on the back of the elephants?

Answer: Most likely tourists
\\~\\
Keep in mind that you should make your question more natural, meaning that the question is plausible to be asked by a human. 
\\~\\
Please generate an unanswerable question and an answerable question for the given caption, in the following format:

Q1: $<$Unanswerable question$>$

A1: $<$answer to Q1$>$

Q2: $<$Answerable question$>$

A2: $<$answer to Q2$>$
\\~\\
DO NOT ask about anything that is difficult to observe or learn even with external knowledge, such as the exact time, exact location, the exact thought of someone, or the conversation or the topic of conversation between people. If you can only come up with such a question, put "Not a good question" for A1.}

\end{tcolorbox}

%%%%%%%%%%

\begin{tcolorbox}[colback=cyan!5!white, colframe=cyan!75!black, title= Complex (Epistemic Awareness), breakable]
\small{You are given a caption of an image. Generate unanswerable questions that asks about an existing object in the image, but is too complex even for humans to answer. 
The unanswerable question should be extremely difficult in framing or tedious to infer the answer.
The answerable question should have a convoluted framing but should have an accurate and direct answer.

Here are some examples:
\\~\\
Caption: This image captures a serene moment in a zoo enclosure, where two majestic giraffes are seen in their natural behavior. The giraffes, adorned in their distinctive brown and white patterns, stand tall against the backdrop of lush green trees. On the left, one giraffe is actively engaged in a meal, its long neck extended towards the tree as it munches on the verdant leaves. Its companion on the right stands leisurely next to a tree trunk, perhaps taking a break from its own leafy feast. The enclosure they inhabit is grassy and spacious, providing them with ample room to roam and forage. The trees dotting the enclosure not only offer a source of food but also create a naturalistic habitat for these towering creatures. In summary, this image is a snapshot of life in a zoo, showcasing the grace and beauty of giraffes in an environment designed to mimic their wild habitats.
\\~\\
Unanswerable Question: How many tree leaves are seen in the image? 

Answer: Unanswerable

Answerable Question: How many animal legs are present?

Answer: 8 legs of 2 girraffes
\\~\\
Caption: This image captures a fascinating scene in a dense jungle. Two majestic, gray elephants are the main subjects of the photo. They are carrying people on their backs, who are seated in wooden seats and wearing helmets for safety. The elephants are walking in a line, one following the other, on a path that cuts through the lush greenery of the jungle. The photo is taken from a higher vantage point, providing a bird's eye view of the elephants and their verdant surroundings. The dense foliage and towering trees of the jungle envelop the path, creating a sense of adventure and exploration.
\\~\\
Unanswerable question: What are the interactions of the individuals on the elephants' backs with the environment?

Answer: Unanswerable

Answerable question: A couple of living beings are carrying another couple of living beings. What are the latter living beings?

Answer: Humans
\\~\\
IMPORTANT: COMPLEXITY OF THE QUESTION SHOULD BE ONLY AND ONLY BASED ON DIFFICULTY TO ANSWER OR FRAMING OF THE QUESTION.
THEY SHOULD NOT REQUIRE ADDITIONAL INFORMATION.
\\~\\
Please generate an unanswerable question and an answerable question for the given caption, in the following format:

Q1: $<$unanswerable question$>$

A1: $<$answer to Q1$>$

Q2: $<$answerable question$>$

A2: $<$answer to Q2$>$ }

\end{tcolorbox}

%%%%%%%%%%

For the extraneous category, we first identify the noun phrases that 
are most relevant to the answer, so that the absence of this object would make it difficult to answer the question. We then mask out the object using Grounded-SAM and inpaint the mask to obtain a perturbed image. Following this, we provide the original and the perturbed image and prompt GPT-4V to generate a question that is answerable for only one of the images.

\begin{tcolorbox}[colback=cyan!5!white, colframe=cyan!75!black, title=Identification of salient objects for extraneous (Epistemic Awareness), breakable]
\small{You are given a question and an answer based on an image. Return the most relevant object in the image that the question is asking about. 
\\~\\
There are some policies to follow:

1. The most relevant object should be the one that when removed from the image, the question would become unanswerable. Here are some examples:

  - {``question": ``What is the color of the car?",  ``answer": ``red"}
  
    Relevant object: red car
  
  - {``question": ``What objects are reflected?",  ``answer": ``trees"} 
    Relevant object: trees
  
  - {``question": ``What brand of bike can you see?",  ``answer": ``yamaha"} 
  
    Relevant object: yamaha bike
  
  - {"question": "What is stopping the animals from running away?",  "answer": "wall"} 
    
    Relevant object: wall
\\~\\
2. Remember that are limitations in removing object from the image. If the question is regarding the overall presentation of the image, it is impossible to masking out the whole image, so the answer should be na. For example,
\\~\\
  - {``question": ``Is this picture taken during the day or night?",  ``answer": ``day"}
  
    Relevant object: na
  
  - {``question": ``Is this a house kitchen or a restaurant kitchen?",  ``answer": "restaurant"}
  
    Relevant object: na
  Don't over do it for policy 2, for example,
  
  - {``question": ``Is the rider a child or an adult?", ``answer": ``adult"}
    Relevant object: adult rider

3. Imagine that even after masking the most relevant object, the question can still be answered, then the answer should be na. For example,
\\~\\
  - {``question": ``What is the woman standing on?",  ``answer": ``floor"}
    
    Relevant object: na
    
    Reasoning: we can still reason that she is standing on the floor, given the rest of the context of the image
  
  - {``question": ``What is the person standing on?",  ``answer": "ski"}
  
    Relevant object: na
    
    Reasoning: we can still reason that he or she is standing on snow, given the rest of the context of the image
\\~\\
4. In the case that there are rich descriptions about the object mentioned in the question, the answer should be the most relevant object that is mentioned in the question, and please try keep the decription intact. For example,
\\~\\
  - {``question": ``What does the sign on the door on the bottom right say?",  ``answer": ``caution"} 
  
    Relevant object: the caution sign on the door on the bottom right
  
  - {``question": ``What stuffed animal is the child in the red jacket holding?",  ``answer": ``teddy bear"} 
  
    Relevant object: teddy bear that the child in the red jacket is holding
\\~\\
5. When the question can be answered, regardless of what is in the image

  - {``question": ``Glasses assist in helping what organ?",  ``answer": ``eyes"} 
    
    Relevant object: na
\\~\\
6. For questions that are general, please evaluate how often there might be multiple objects belonging to the same category appearing in a scene, and return the most plausible answer. For example,
\\~\\  
  - {``question": ``What food is presented?",  ``answer": "sandwich"} 
    Relevant object: ``food"
  
  - {``question": ``What is being eaten?",  ``answer": ``sandwich"} 
    Relevant object: "food"}

\end{tcolorbox}

%%%%%%%%%%

\begin{tcolorbox}[colback=cyan!5!white, colframe=cyan!75!black, title=Prompt to generate Extraneous category (Epistemic Awareness)]
\small{You are given a pair of very similar images. In image 2, there is a specific object that is missing or changed from image 1. Generate a question that is answerable for image 1 while not answerable for image 2.
\\~\\
There are a few rules to follow for each question:

1. The question should be answerable for image 1, that is there is a definitive answer to the question, just by looking at image 1.
\\~\\
2. The question should not be answerable for image 2. "Not answerable" means, just by looking at image 2, the answer would be something like ``I don't know", ``I don't see SOMETHING" or ``Nothing". For example,
    
    - If the question is ``What color is the car?", and there is no car in image 2, the answer should be ``I don't see a car".
    
    - If the question is ``What is on the man's head", and there is nothing on the man's head in image 2, the answer should be ``Nothing".
    
    - If the question is asking about something that cannot be seen clearly in image 2, the answer should be ``I don't know".
    
    - Try not to ask questions about the presence of an object, but rather about the properties of the object. For example, instead of asking ``Is there a car in the image?", ask ``What color is the car?". Instead of asking ``How many people are there?", ask "What is the person wearing?".
\\~\\   
3. The question should be relevant to the content of each image alone, even without seeing the other image.
\\~\\
The response should be formatted as:

- Q: $<$question$>$

- A1: $<$answer for image 1$>$

- A2: $<$answer for image 2, choose your answer from ``I don't know", ``I don't see xxx" or ``Nothing". Try not to refer to the answer for image 1$>$}

\end{tcolorbox}

%%%%%%%%%%

\begin{tcolorbox}[colback=cyan!5!white, colframe=cyan!75!black, title=Ambiguous (Aleotoric Awareness), breakable]
\small{You are given a caption of an image. Generate unanswerable questions that asks about an existing object in the caption, but is ambiguous. 

DEFINITION: Ambiguity refers to a situation or statement that can be understood or interpreted in multiple ways. It often involves uncertainty or lack of clarity, leading to confusion or different possible meanings.

The unanswerable question should be ambiguous because of indifferentiablity of objects or people mentioned in the question. As a result without clarification, multiple answers are possible.
The answerable question should have a convoluted framing but should have an accurate and direct answer.

Here are some examples:
\\~\\
Caption: This image captures a serene moment in a zoo enclosure, where two majestic giraffes are seen in their natural behavior. The giraffes, adorned in their distinctive brown and white patterns, stand tall against the backdrop of lush green trees. On the left, one giraffe is actively engaged in a meal, its long neck extended towards the tree as it munches on the verdant leaves. Its companion on the right stands leisurely next to a tree trunk, perhaps taking a break from its own leafy feast. The enclosure they inhabit is grassy and spacious, providing them with ample room to roam and forage. The trees dotting the enclosure not only offer a source of food but also create a naturalistic habitat for these towering creatures.In summary, this image is a snapshot of life in a zoo, showcasing the grace and beauty of giraffes in an environment designed to mimic their wild habitats.
\\~\\
Unanswerable Question: What is the giraffe doing?

Answer: There are multiple giraffes. Unanswerable

Answerable Question: Where are the people sitting?

Answer: All people are sitting on elephants' backs.
\\~\\
Caption: This image captures a fascinating scene in a dense jungle. Two majestic, gray elephants are the main subjects of the photo. They are carrying people on their backs, who are seated in wooden seats and wearing helmets for safety. The elephants are walking in a line, one following the other, on a path that cuts through the lush greenery of the jungle. The photo is taken from a higher vantage point, providing a bird's eye view of the elephants and their verdant surroundings. The dense foliage and towering trees of the jungle envelop the path, creating a sense of adventure and exploration.
\\~\\
Unanswerable question: Is the bird's eye view from the top of a tree or from a nearby mountain or a drone?

Answer: All options are possible. Unanswerable

Answerable question: What are the people on the elephants' backs wearing?

Answer: Helmets
\\~\\
IMPORTANT: AMBIGUITY OF THE QUESTION SHOULD BE ONLY AND ONLY BASED ON THE POSSIBILITY OF MULTIPLE ANSWERS.
THEY SHOULD NOT REQUIRE ADDITIONAL INFORMATION.
\\~\\
Please generate an unanswerable question and an answerable question for the given caption, in the following format:

Q1: $<$unanswerable question$>$

A1: answer to Q1

Q2: $<$answerable question$>$

A2: answer to Q2}

\end{tcolorbox}

%%%%%%%%%%

%%%%%%%%%%

\begin{tcolorbox}[colback=cyan!5!white, colframe=cyan!75!black, title=Temporal (Aleatoric Awareness), breakable]
\small{You are given a caption of an image. Generate a question that requires to make predictions of future events from the time the image is captured requiring some temporal event reasoning that is not directly observable from the image. 
An unanswerable question requires temporal reasoning that cannot be inferred from the caption to answer the question.
An answerable question requires temporal commonsense and can be answered from the caption.

Make the unanswerable and answerable questions as similar to each other as possible yet one is answerable and the other is unanswerable. 
Do NOT ask about anything that is difficult to infer even if you observe the future events, such as the exact time, exact location, or the exact thought of someone.

Here are some examples:
\\~\\
Caption: The image showcases a captivating scene of a dressage routine being performed by two horses and their riders in a grassy field. The horse on the left is a majestic white stallion, while the one on the right is a striking black stallion. Both horses are displaying their strength and agility by rearing up on their hind legs, creating an impressive spectacle.The riders, dressed in crisp white outfits and blue hats, appear to be in perfect sync with their horses. Their attire contrasts beautifully with the vibrant green of the field, adding to the overall aesthetic of the image.In the background, colorful flags and obstacles can be seen, indicating that this might be a competitive event. The lush trees and shrubs further enhance the natural beauty of the setting.Overall, this image captures a moment of harmony between the riders and their horses, set against a backdrop of nature's splendor. It's a testament to the skill and grace involved in dressage.
\\~\\
Unanswerable Question: Are the two people riding the horses going to fall?

Answer: Unanswerbale

Answerable Question: Has the race started?

Answer: Yes
\\~\\
Caption: The image features two main objects placed on a white shelf against a white wall. On the left, there is a charming **owl candle holder**. It is white in color, matching the overall aesthetic of the setting. The owl's intricate design is captivating, with its wide eyes and detailed feathers. A candle is lit within the holder, casting a warm and inviting glow.To the right of the owl, there is an **antique-style clock**. The clock is predominantly white but is adorned with gold accents that add a touch of elegance. It has a round face displaying time with Roman numerals, further enhancing its antique appeal.The objects are arranged neatly on the shelf, creating a harmonious and pleasing visual composition. The use of white for both the objects and the background gives the image a clean, minimalist aesthetic.
\\~\\
Unanswerable Question: Is the clock still working?

Answer: Unanswerable

Answerable Question: Is someone lighting the candle?

Answer: No, it is already lit.
\\~\\
Keep in mind that you should make your question more natural, meaning that the question is plausible to be asked by a human. 
\\~\\
IMPORTANT: The question and answer should only have temporal aspect.

IMPORTANT: The answerable question should have a clear and correct answer without any ambiguity.
\\~\\
Please generate an unanswerable question and an answerable question for the given caption, in the following format:

Q1: $<$unanswerable question$>$

A1: $<$answer to Q1$>$

Q2: $<$answerable question$>$

A2: $<$answer to Q2$>$}

\end{tcolorbox}

%%%%%%%%%%

\subsection{Prompts for Evaluation}
Here are the prompts for evaluation:

\begin{tcolorbox}[colback=cyan!5!white, colframe=cyan!75!black, title=LAVE Refusal detection, breakable]
\small{You are given a question and an answer. Please decide whether the answer is a refusal to the question. 
    
Give a rating of 0 or 1. 1 indicates a refusal, 0 indicates not a refusal. Give the rationale after rating.
\\~\\
Here are a few examples you can consider:
\\~\\
Question: ``What is the capital of France?''

Answer: ``I don't know''

Rating: 1

Rationale: The candidate answer is a refusal because it does not provide an answer to the question. Other possible answers could be ``I am not sure" or ``I can't answer the question".
\\~\\
Question: ``What is the color of the car?''

Answer: ``I don't see a car."

Rating: 1

Rationale: The candidate answer is a refusal because it does not provide an answer to the question. Other possible answers could be ``There is no car" or ``I can't see the car".
\\~\\
Question: ``What is the person holding?"

Answer: ``Nothing"

Rating: 1

Rationale: The candidate answer is a refusal because the answer is simply ``nothing". Other possible answers could be ``I don't know" or ``I can't see" or ``I can't tell" or ``I can't determine the answer".
\\~\\
Question: ``What is the color of the car?"

Answer: ``It is not possible to tell the color of the car from the image. However, cars are commonly in the following colors, ...."

Rating: 1

Rationale: The candidate answer is a refusal because the answer clearly states that it is not possible to give an definite answer. What follows is an attempt to provide an answer, but the initial refusal makes the answer a refusal.
\\~\\
Question: ``What is the color of the car?"

Answer: ``I am not sure, but I think the answer is blue"

Rating: 0

Rationale: The candidate answer provides an answer to the question.
\\~\\
Question: ``What is the person holding?"

Answer: ``an apple, but I am not confident"

Rating: 0

Rationale: The candidate answer provides an answer to the question.
\\~\\
Question: ``What is the man wearing?"

Answer: ``he man is wearing a hat that looks like a pineapple."

Rating: 0

Rationale: The candidate answer is not a refusal because it provides an answer to the question.}

\end{tcolorbox}

%%%%%%%%%%

\begin{tcolorbox}[colback=cyan!5!white, colframe=cyan!75!black, title=LAVE accuracy, breakable]
\small{You are given a question, a gold-standard reference answers written by experts, and a candidate answer. Please rate the accuracy of the candidate answer for the question considering the reference answer. Use a scale of 1-3, with 1 indicating an incorrect or irrelevant answer, 2 indicating an ambiguous or incomplete answer, and 3 indicating a correct answer. Give the rationale after rating.
\\~\\
Please follow the following format:
\\~\\
Rating: 1

Rationale: The candidate answer is incorrect because ...}

\end{tcolorbox}

%%%%%%%%%%

% \lj{setting for Grounded SAM, Lama inpainting}

% \textbf{Thresholding.} For Thresholding baselines, we grid search among $(0.1, 0.2, ... 0.9)$ and $(0.91, 0.92, ... 0.99)$ to decide the optimal threshold for each split. The latter range is included, as we observe that the models are often over-confident in its own predictions, which also reflects in ECE in Table~\ref{tab:finetune}.

\end{document}